\documentclass[twoside]{article}

\usepackage[accepted]{aistats2025}
\usepackage[pagebackref,breaklinks,colorlinks]{hyperref}
\hypersetup{
    colorlinks=true,
    linkcolor= blue,
    filecolor=violet,      
    urlcolor= blue,
     citecolor = blue, %blue, %magenta,
    pdfpagemode=FullScreen,
}
\usepackage[noabbrev,capitalize,nameinlink]{cleveref}
\usepackage[authoryear, compress, round]{natbib}
\setcitestyle{nosort}

\usepackage{fontawesome}

\usepackage{import}

\usepackage{mathtools}
\usepackage{enumitem}
\usepackage{bm}
\usepackage{cancel}

\usepackage{amsmath}
\usepackage{amsthm}
\usepackage{amssymb}
\usepackage{pifont}% http://ctan.org/pkg/pifont
\usepackage{algpseudocode}

\newcommand{\cmark}{\ding{51}}%
\newcommand{\xmark}{\ding{55}}%

\newcommand{\eq}{\begin{equation*}}
\newcommand{\en}{\end{equation*}}
\newcommand{\eqa}{\begin{eqnarray*}}
	\newcommand{\ena}{\end{eqnarray*}}
\newcommand{\eqn}{\begin{equation}}
\newcommand{\enn}{\end{equation}}
\newcommand{\be}{\begin{equation}}
\newcommand{\ee}{\end{equation}}
\newcommand{\eqan}{\begin{eqnarray}}
\newcommand{\enan}{\end{eqnarray}}

\newcommand{\nn}{\nonumber}

\newcommand{\bv}[1]{\mathbf{#1}} % bold vector 

\newcommand{\bb}{ {\bf b} }
\newcommand{\ba}{\textbf{a}}

\newcommand{\bi}{ {\bf b} }

\newcommand{\argmax}{\mathop{\mathrm{argmax}}}

\newcommand{\ie}{i.e., }

\newcommand{\eg}{e.g., }

\usepackage{graphicx}
\usepackage{xspace}
\usepackage{bbm}
\usepackage{bm}

\usepackage{diagbox}

\usepackage{enumitem}

\usepackage{color,soul}
\usepackage{xcolor}  

\usepackage{tikz}

\usepackage{pgfplots}
\pgfplotsset{compat=1.18}
\usetikzlibrary{3d,decorations.text,shapes.arrows,positioning,fit,backgrounds}
\usetikzlibrary{intersections, pgfplots.fillbetween}

\definecolor{lineBlue}{RGB}{57,106,177}
\definecolor{lineOrange}{RGB}{218,124,48}
\definecolor{lineGreen}{RGB}{62,150,81}
\definecolor{lineRed}{RGB}{204,37,41}
\definecolor{lineGray}{RGB}{83,81,84}
\definecolor{linePurple}{RGB}{107,76,154}
\definecolor{lineMaroon}{RGB}{146,36,40}
\definecolor{barBlue}{RGB}{114,147,203}
\definecolor{barOrange}{RGB}{225,151,76}
\definecolor{barGreen}{RGB}{132,186,91}
\definecolor{barRed}{RGB}{211,94,96}
\definecolor{barGray}{RGB}{128,133,133}
\definecolor{barPurple}{RGB}{144,103,167}
\definecolor{barMaroon}{RGB}{171,104,81}

\usepackage{multirow}

\usetikzlibrary{arrows.meta}

\usepackage{algorithm}
\usepackage{algpseudocode}
\usepackage{booktabs}

\subimport{./}{macros}

\begin{document}


\runningauthor{Irit Chelly, Roy Uziel, Oren Freifeld, Ari Pakman}

\twocolumn[

\aistatstitle{Consistent Amortized Clustering via Generative Flow Networks}

\aistatsauthor{ Irit Chelly${}^{1}$  \And Roy Uziel${}^{1}$ \And  Oren Freifeld${}^{1,3}$
\And Ari Pakman${}^{2,3}$} 

% \aistatsaddress{ 
% Department of 
% \\Computer Science 
% \And  
% Department of 
% \\Computer Science 
% \\ 
% \\
% \qquad\qquad\qquad\qquad\qquad\qquad\qquad
% Ben-Gurion University of the Negev,
% Beer Sheva, Israel
% \And 
% Department of 
% \\Computer Science 
% \And 
% Department of Industrial \\
% Engineering and Management,
% \\
% The School of Brain Sciences and Cognition
% } ]

\aistatsaddress{ 
\qquad \qquad \qquad \qquad \qquad \qquad \qquad \qquad \qquad \qquad 
\qquad \qquad \qquad \qquad \qquad \qquad \qquad \qquad \qquad 
${}^{1}$Department of  Computer Science 
\qquad 
${}^{2}$Department of Industrial Engineering and Management
\And 
\\
\qquad\qquad\qquad\qquad\qquad\qquad\qquad
${}^{3}$The School of Brain Sciences and Cognition
\\
\qquad\qquad\qquad\qquad\qquad\qquad\qquad
Ben-Gurion University of the Negev,
Beer Sheva, Israel
\And 
\And 
} ]

\begin{abstract}

Neural models for amortized probabilistic clustering yield samples of cluster labels given a set-structured input, while avoiding lengthy Markov chain runs and the need for explicit data likelihoods.
Existing methods which label each data point 
sequentially, like the Neural Clustering Process, often lead to cluster assignments highly dependent on the data order.
Alternatively, methods that sequentially create full clusters, 
do not provide assignment probabilities.
In this paper, we introduce GFNCP, a novel framework for  amortized clustering.
GFNCP is formulated as a Generative Flow Network with a shared energy-based parametrization of policy and reward.
We show that the flow matching conditions are equivalent to consistency of the clustering posterior under marginalization, which in turn implies order invariance. 
GFNCP also outperforms existing methods in clustering performance on both synthetic and real-world data.
%Our code will be made publicly available upon acceptance.

\end{abstract}

\section{INTRODUCTION}\label{sec:intro}

Probabilistic clustering models, also known as mixture models, play a crucial role in 
many scientific domains and are extensively used in various downstream tasks. These models aim to learn the underlying data structure by grouping similar data points into clusters, with cluster assignments encoded in  
the posterior distribution of  discrete latent variables.

\begin{figure}[hbt!]
    % \renewcommand{\arraystretch}{0.04}
    % \centering
    \Large
    \begin{tabular}{p{3.6cm}p{3.6cm}}
    \scalebox{0.495}{
    \hspace{-1.4cm}
        % This file was created with tikzplotlib v0.10.1.
\begin{tikzpicture}

\definecolor{darkgray176}{RGB}{176,176,176}
\definecolor{green_1}{RGB}{173,216,230}
% \definecolor{green_1}{RGB}{173,216,230}
\definecolor{green_1}{rgb}{0.6, 0.81, 0.93}
\definecolor{forestgreen4416044}{RGB}{44,160,44}
\definecolor{red_1}{RGB}{178,24,24}
\definecolor{green_1}{rgb}{0.4, 0.69, 0.2}

\begin{axis}[
tick align=outside,
tick pos=left,
x grid style={darkgray176},
xlabel={SDPP Metric},
xmin=-0.13, xmax=10,
xtick style={color=black},
y grid style={darkgray176},
ylabel={Frequency},
ymin=0, ymax=320,
ytick style={color=black},
y label style={at={(axis description cs:-0.16,.5)},anchor=south},
x label style={at={(axis description cs:.5,-0.12)}},
]

\node[text width=1cm] at (6.4,290) {\large GFNCP};
\node[text width=1cm] at (6.9,265) {\large NCP};
\draw[draw=black,fill=green_1] (axis cs:8,280) rectangle (axis cs:9.5,300);
\draw[draw=black,fill=red_1,opacity=0.4] (axis cs:8,255) rectangle (axis cs:9.5,275);
\draw[draw=black,fill=green_1,fill opacity=0.0] (axis cs:5.5,249) rectangle (axis cs:9.7,306);

\draw[draw=black,fill=green_1] (axis cs:0.12332564212824,0) rectangle (axis cs:0.385047029624798,63);
\draw[draw=black,fill=green_1] (axis cs:0.385047029624798,0) rectangle (axis cs:0.646768417121357,208);
\draw[draw=black,fill=green_1] (axis cs:0.646768417121357,0) rectangle (axis cs:0.908489804617915,224);
\draw[draw=black,fill=green_1] (axis cs:0.908489804617915,0) rectangle (axis cs:1.17021119211447,144);
\draw[draw=black,fill=green_1] (axis cs:1.17021119211447,0) rectangle (axis cs:1.43193257961103,105);
\draw[draw=black,fill=green_1] (axis cs:1.43193257961103,0) rectangle (axis cs:1.69365396710759,69);
\draw[draw=black,fill=green_1] (axis cs:1.69365396710759,0) rectangle (axis cs:1.95537535460415,53);
\draw[draw=black,fill=green_1] (axis cs:1.95537535460415,0) rectangle (axis cs:2.21709674210071,40);
\draw[draw=black,fill=green_1] (axis cs:2.21709674210071,0) rectangle (axis cs:2.47881812959726,13);
\draw[draw=black,fill=green_1] (axis cs:2.47881812959726,0) rectangle (axis cs:2.74053951709382,15);
\draw[draw=black,fill=green_1] (axis cs:2.74053951709382,0) rectangle (axis cs:3.00226090459038,15);
\draw[draw=black,fill=green_1] (axis cs:3.00226090459038,0) rectangle (axis cs:3.26398229208694,15);
\draw[draw=black,fill=green_1] (axis cs:3.26398229208694,0) rectangle (axis cs:3.5257036795835,10);
\draw[draw=black,fill=green_1] (axis cs:3.5257036795835,0) rectangle (axis cs:3.78742506708005,6);
\draw[draw=black,fill=green_1] (axis cs:3.78742506708006,0) rectangle (axis cs:4.04914645457661,9);
\draw[draw=black,fill=green_1] (axis cs:4.04914645457661,0) rectangle (axis cs:4.31086784207317,2);
\draw[draw=black,fill=green_1] (axis cs:4.31086784207317,0) rectangle (axis cs:4.57258922956973,3);
\draw[draw=black,fill=green_1] (axis cs:4.57258922956973,0) rectangle (axis cs:4.83431061706629,2);
\draw[draw=black,fill=green_1] (axis cs:4.83431061706629,0) rectangle (axis cs:5.09603200456284,3);
\draw[draw=black,fill=green_1] (axis cs:5.09603200456284,0) rectangle (axis cs:5.3577533920594,1);

\draw[draw=black,fill=red_1,opacity=0.4] (axis cs:0.0244094305721065,0) rectangle (axis cs:0.751850527804012,183);
\draw[draw=black,fill=red_1,opacity=0.4] (axis cs:0.751850527804012,0) rectangle (axis cs:1.47929162503592,302);
\draw[draw=black,fill=red_1,opacity=0.4] (axis cs:1.47929162503592,0) rectangle (axis cs:2.20673272226782,172);
\draw[draw=black,fill=red_1,opacity=0.4] (axis cs:2.20673272226782,0) rectangle (axis cs:2.93417381949973,121);
\draw[draw=black,fill=red_1,opacity=0.4] (axis cs:2.93417381949973,0) rectangle (axis cs:3.66161491673163,74);
\draw[draw=black,fill=red_1,opacity=0.4] (axis cs:3.66161491673163,0) rectangle (axis cs:4.38905601396354,44);
\draw[draw=black,fill=red_1,opacity=0.4] (axis cs:4.38905601396354,0) rectangle (axis cs:5.11649711119544,31);
\draw[draw=black,fill=red_1,opacity=0.4] (axis cs:5.11649711119544,0) rectangle (axis cs:5.84393820842735,24);
\draw[draw=black,fill=red_1,opacity=0.4] (axis cs:5.84393820842735,0) rectangle (axis cs:6.57137930565925,12);
\draw[draw=black,fill=red_1,opacity=0.4] (axis cs:6.57137930565925,0) rectangle (axis cs:7.29882040289116,13);
\draw[draw=black,fill=red_1,opacity=0.4] (axis cs:7.29882040289116,0) rectangle (axis cs:8.02626150012306,5);
\draw[draw=black,fill=red_1,opacity=0.4] (axis cs:8.02626150012307,0) rectangle (axis cs:8.75370259735497,4);
\draw[draw=black,fill=red_1,opacity=0.4] (axis cs:8.75370259735497,0) rectangle (axis cs:9.48114369458687,4);
\draw[draw=black,fill=red_1,opacity=0.4] (axis cs:9.48114369458687,0) rectangle (axis cs:10.2085847918188,2);
\draw[draw=black,fill=red_1,opacity=0.4] (axis cs:10.2085847918188,0) rectangle (axis cs:10.9360258890507,5);
\draw[draw=black,fill=red_1,opacity=0.4] (axis cs:10.9360258890507,0) rectangle (axis cs:11.6634669862826,1);
\draw[draw=black,fill=red_1,opacity=0.4] (axis cs:11.6634669862826,0) rectangle (axis cs:12.3909080835145,1);
\draw[draw=black,fill=red_1,opacity=0.4] (axis cs:12.3909080835145,0) rectangle (axis cs:13.1183491807464,0);
\draw[draw=black,fill=red_1,opacity=0.4] (axis cs:13.1183491807464,0) rectangle (axis cs:13.8457902779783,1);
\draw[draw=black,fill=red_1,opacity=0.4] (axis cs:13.8457902779783,0) rectangle (axis cs:14.5732313752102,1);
\end{axis}

\end{tikzpicture}
    } & 
    \scalebox{0.495}{
        \hspace{-0.7cm}
        \input{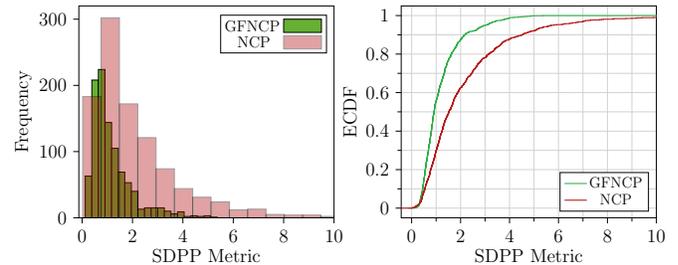}
    }
    \end{tabular}
    \caption{Histogram and Empirical CDF of the SDPP metric (\autoref{eq:inv_metric}) computed on GFNCP's and NCP's clustering results, trained on MNIST dataset. GFNCP shows a substantial improvement in producing consistent samples across different data orders, as it puts more probability on low SDPP values.}    
    \label{fig:invariance_ecdf_mnist}
\end{figure}

Given a generative model for the clusters, traditional posterior inference methods, such as Markov Chain Monte Carlo (MCMC) \citep{neal:2000:MCMC} and their fast implementations~\citep{Chang:NIPS:2013:ParallelSamplerDP, dinari:2019:distributed, dinari2022cpu}, yield samples from the posterior distribution, providing an asymptotically exact solution given enough samples. However, they often struggle with high-dimensional data or large datasets. Conversely, variational inference \citep{Blei:2004:VMD} is more suitable for large datasets; however, it incurs a trade-off in accuracy as a result of its reliance on approximations. 
%
% Deep learning methods: 

Over the last decade, with the increasing intricacy of data, a variety of clustering solutions have been proposed that utilize deep neural models. 
A popular family of models, reviewed in~\cite{ren2024deep} and~\cite{zhou:2022:survey1}, and often referred to as performing {\it deep clustering}, are  unsupervised classifiers trained to discover a finite number of categories. Among these models, methods such as DCN \citep{yang:2017:DCN} and ClusterGAN \citep{mukherjee:2019:clstrGAN} are designed to simultaneously tackle both data representation and clustering tasks.
Another approach involves a two-step process in which feature learning and clustering are decoupled, with SCAN \citep{van:2020:SCAN} and DDC \citep{ren:2020:DDC} being notable examples. 
Importantly, at test time, these models are limited to generating (soft) assignments for {\it individual} data points, since they are restricted by their reliance
on pre-learned categories and their inability to model point interactions. Note that this is also the case for DeepDPM~\citep{ronen:2022:DeepDPM}, despite its ability to infer the number 
of clusters present in the training (but not test) data. 

Our focus in this paper is on a different category of models, which 
address the more ambitious task of {\it jointly} modeling the 
clustering posterior for set-structured data of arbitrary size. 
This task is more challenging, 
as it involves not only learning the structure of individual data points but also the correlation structure among the points of a dataset of any size.
Motivated by this need, there has been consistent progress 
in recent years  in amortized inference methods within the framework of probabilistic clustering, accommodating a set-structured input.
%
% Amortized inference~\citep{Gershman:2014:amortized} refers to the process of training a neural network that, given observations drawn from a generative model containing latent variables, can infer their posterior distribution to produce new observations. 
%
Amortized inference~\citep{Gershman:2014:amortized} refers to the process of training a neural network that can infer the posterior distribution of latent variables based on observations drawn from a generative model.
In this setting, \cite{lee2019set} introduced the Set Transformer~(ST), an attention-based network specifically designed to model point interactions within a set. 
This architecture was used to amortize the inference 
over the {\it parameters} of a mixture model, but was restricted to  
a fixed number of Gaussian components. 
To avoid these limitations, other models directly amortize the posterior over the joint clustering {\it labels} of the dataset.
DAC~\citep{DAC} and CCP~\citep{pakman2020} build on this concept by sequentially generating full clusters, 
%without imposing a fixed number of clusters or a likelihood model, 
thus enabling more complex prior distributions.
%(see also \cite{liu:2021:CHiGAC}, \cite{liu:2022:AMCP}). 
% in which the proposed model employs two strategies to explore relationships among data points.
%
% A disadvantage of these methods is that they do not produce probabilities for the clustering configurations they generate.
%
% This drawback is addressed by an

An alternative approach, named Neural Clustering Process (NCP)~\citep{pakman2020}, sequentially assigns cluster labels to data points.
Unlike previous methods, NCP provides assignment probabilities, thereby offering deeper insights into clustering outputs through uncertainty quantification.
However, these probabilities are highly sensitive to data-order permutations, failing to preserve a fundamental symmetry of the posterior distribution.
The limitations of existing work, as outlined, motivate our current research.

% Our model
In this paper we propose the GFlowNet-based Clustering Process (GFNCP), a posterior generative clustering model that, given a set-structured input sampled from a (possibly infinite) mixture model, yields clustering assignment samples along with their associated probabilities.
%
% In GFNCP, we exploit the framework of GFlowNet to model the clustering problem as a generative process of forming assignment estimations, which are interpreted as final-state objects, and apply marginal consistency flaw
%
In GFNCP, we exploit the framework of GFlowNets~\citep{bengio2021flow} to model the clustering task as a sequential generative process in which the final object, a full-data assignment, is constructed by sampling from a learned policy at intermediate states. 
Notably, we formulate the policy and the learned rewards  as energy-based models with shared parameters in an end-to-end framework.
Our work differs from previous approaches in several aspects (see \autoref{tab:checklist}); in particular, unlike NCP, our model encourages invariance under data order permutations (see \autoref{fig:invariance_ecdf_mnist}).

\begin{table}[hbt!]  %[ht]
    \centering
    \setlength{\abovetopsep}{1.3ex}
    \caption{Amortized-clustering approaches.}
    \resizebox{1\linewidth}{!}{ 
    \begin{tabular}{p{3.45cm}>{\centering\arraybackslash}p{0.3cm}>{\centering\arraybackslash}p{0.7cm}>{\centering\arraybackslash}p{0.7cm}>
    {\centering\arraybackslash}p{0.7cm}>{\centering\arraybackslash}p{0.98cm}}
         \toprule
         \footnotesize PROPERTY & \footnotesize ST & \footnotesize DAC & \footnotesize CCP &\footnotesize NCP & \footnotesize GFNCP \\
         \midrule
         Data-perm. invariance & \cmark & \cmark & \xmark &  \xmark & \cmark \\
         Unlimited components & \xmark & \cmark & \cmark & \cmark & \cmark \\
         Arbitrary likelihood & \xmark & \cmark & \cmark & \cmark & \cmark \\
         Assignment prob. & \xmark & \xmark & \xmark & \cmark & \cmark \\
         Well defined posterior & -- & \xmark & \cmark & \cmark & \cmark \\
         \bottomrule
    \end{tabular}
    }
    \label{tab:checklist}
\end{table}

% Contributions:
Overall, \textbf{our contributions are as follows}: (1)~We present GFNCP, an amortized clustering method formulated as a GFlowNet sequential generative model, which uses an energy-based joint policy and reward function, allowing an unlimited number of components; 
(2) We demonstrate that GFNCP surpasses existing methods in clustering performance and also generalizes better to unseen classes; 
%
%(2) We demonstrate that GFNCP not only surpasses existing methods in clustering performance but also supports an unlimited number of components and generates assignment probabilities. 
%
(3)~We show that GFNCP exhibits greater consistency across different data orders; 
(4) Unlike previous works on cluster label amortization, we show that training can be performed without true labels via instance discrimination.\footnote{Our code is available at  
\url{https://github.com/BGU-CS-VIL/GFNCP}.
}

% Overall, \textbf{our contributions are as follows}: (1) We diagnose a consistency flaw in previous approaches to point-wise amortized clustering models. (2) We formulate amortized clustering as a sequential generative model via GFlow networks with an energy-based policy function. (3) We show that point-wise amortized clustering can be trained without known labels.

\section{RELATED WORK}\label{sec:related_work}

\paragraph{Amortizing discrete variables.}
Beyond clustering, models exist for posteriors over  
permutations~\citep{mena2018learning,pakman2020neural} 
and network communities~\citep{wang2024amortized}, {\it inter alia.} 
When both a generative and an inference model over discrete latents are learned, reparametrization gradients can be used via continuous relaxations~\citep{maddison2017concrete, jang2017categorical}. 
To amortize distributions 
over discrete factor graphs, \cite{buesing2020approximate} 
formulate sampling  as a MaxEnt Markov Decision Process. However, this approach fails when there are many ways to generate the same object, as in our setting.
Other approaches that leverage nonparametric Bayesian models within neural networks, such as those proposed in \citep{nalisnick2016stick} and \citep{jiang2016variational}, develop generative models with latent discrete labels.

%%%%%%%%%%%%%%%%%%%%%%%%%%%%%%%%%%%%%%%%%%%%%%%%%%%%%%%%%%%%%%%%%%%%%%%%

\paragraph{Generative Flow Networks (GFlowNets).} Introduced in \citep{bengio2021flow} and 
reviewed in~\autoref{sec:gflonet_review}, this set of algorithms is designed to train a stochastic policy for sampling composite objects from a target distribution, following a sequence of actions structured as a directed acyclic graph (DAG).
GFlowNets address the challenging setting in which different trajectories in the space of actions can yield the same final state.
%
% was first introduced as reinforcement-learning algorithm aimed at learning a stochastic policy for generating an object (final state) through a sequence of actions, with the probability proportional to a specified reward function. GFlowNet aims to solve the challenging setting in which different trajectories can yield the same final state.
%
GFlowNets have connections to variational inference~\citep{malkin:2022:GFlowNets_VI} and entropy-regularized reinforcemnt learning~\citep{tiapkin2024generative, deleudiscrete}, and have been applied to problems 
such as biological and natural language sequences~\citep{jain2022biological,huamortizing} and  combinatorial optimization~\citep{zhang2023let}. 
Models similar to ours, were GFlowNets are conditioned
on data, include~\cite{deleu2022bayesian,hu2023gflownet}. 
%
%
% \cite{zhang2022generative} amortizes the MCMC algorithm by introducing a framework that jointly trains a GFlowNet with an energy function, which serves as a learned reward function, across two different models.

\section{BACKGROUND}
\subsection{Generative  models of clusters} 
Consider  $N$ data points $\mathbf{ x} = \{x_i\}$, 
and assume they were generated through a probabilistic clustering model of the form
\begin{align}
\nn
\alpha_1, \alpha_2 &\sim p(\alpha_1, \alpha_2)
\\
\nn 
N &\sim p(N)
\\
c_1 \ldots c_N &\sim p(c_{1:N}|\alpha_1) 
\quad c_i \in \{ 1 \ldots K \}
\label{eq:gen1}
\\
\mu_1 \ldots \mu_K|c_{1:N} &\sim 
p(\mu_{1:K}  |\alpha_2) 
\nn
\\
x_i &\sim p(x_i|\mu_{c_{i}}) \quad i=1 \ldots N.
\nn
\end{align}
%The generative model introduces discrete random variables~$c_i$ denoting the index of the cluster that data point~$x_i$ is assigned to,
% and note that $K$, the number of clusters, can itself be a random variable. 
%
The generative model introduces discrete random variables~$c_i$, representing the cluster index for each data point~$x_i$. Note that $K$, the number of clusters,
can itself be a random variable. 
Here $\alpha_1, \alpha_2$ are hyperparameters and~$\mu_k$ denotes a parameter vector controlling the distribution of the $k$-th cluster (e.g.,~$\mu_k$ could include both the mean and covariance of a Gaussian-mixture component). 
We also assume that the priors $p(c_{1:N}|\alpha_1)$ and 
$p(\mu_{1:K}|\alpha_2)$ are exchangeable,
\eqan 
\nn
p(c_{1:N}|\alpha_1) &=& 
p(c_{\rho_1}\ldots c_{\rho_N}|\alpha_1)\,,
\\
p(\mu_{1:K} |\alpha_2) &=& 
p(\mu_{\xi_1} \ldots \mu_{\xi_K}|\alpha_2)\,,
\label{eq:exch_pior}
\enan
where $\{ \rho_i \}_{i=1}^N$ and $\{ \xi_k \}_{k=1}^K$ are arbitrary permutations over $N$ and $K$ 
indices, respectively.

% \begin{figure*}[t!]
% 	\begin{center}
% 		\fbox{		
% 		\includegraphics[width=.85\textwidth]{figures/Gks2.pdf}}
% 	\end{center}
% 	\caption{{\bf Encoding cluster labels.} 
% After assigning labels 
% $c_{1:6}$ to $K=2$ clusters, each of the three possible $c_7$
% labels  (for the circled point~$x_7$) gives
% an encoding  $G_k$ for the set $x_{1:7}$.  
% Each $G_k$ is the encoding $G_{c_{1:6}}$ obtained when 
% point $x_n$ joins cluster $k$. 
% The vector~$U= U_{n+1}$ encodes the four gray unlabeled points (Best in color).
% 	} 
% 	\label{fig:Gks}	
% \end{figure*}
Given data points $x_{1:N}$, the central object we aim to model 
is the posterior $p(c_{1:N}|x_{1:N})$. 
Of particular interest for us are two properties: 
% posterior Note in  As in many distributions over highly structured domains,  
%  typically respect  certain symmetries~\citep{kallenberg2005probabilistic}.
% \paragraph{Labeling Invariance}
% It should not matter which label we use to designate each cluster: 
% \eqan 
% p(c_{1:N} |x_{1:N} ) = 
% p(\xi_{c_{1}} \ldots \xi_{c_N}|x_{1:N}) \,.
% \label{eq:label_sym}
% \enan 
\begin{itemize}
    \item {\bf Conditional exchangeability:}
probabilities should not 
depend on the order of the data: 
% As follows from (\ref{eq:gen1}) and (\ref{eq:exch_pior}), we have
\begin{equation}
p(c_{1:N} |x_{1:N}) = 
p(c_{\rho_1}\ldots c_{\rho_N}|x_{\rho_1} \ldots x_{\rho_N} )\,.
\label{eq:exchange}   
\end{equation}

\item {\bf Marginal consistency:}
by definition, the following relationship holds between marginals
(for $n=1 \ldots N-1$): 
\eqan 
p(c_{1:n}| x_{1:N}  )  = 
 \sum_{c_{n+1}} p(c_{1:n}, c_{n+1}| x_{1:N} ) \,. 
 \label{eq:marginal_consistency1}
\enan 
\end{itemize}

\subsection{The Neural Clustering Processes}
\label{sec:amortizing}
The standard approach to draw samples from the posterior $p(c_{1:N}|x_{1:N})$, MCMC, has two major limitations. First, convergence can be slow and hard to assess. Second, MCMC requires an explicit expression for the generative model $p(x_i|\mu_{c_i})$. 
A common way out of the latter problem is to assume that $p(x_i|\mu_{c_i})$ follows a Gaussian
(perhaps after some data pre-processing, \eg, 
as proposed in~\cite{Chang:NIPS:2013:ParallelSamplerDP}),
%~\citep{dinari:2022:sampling,ronen:2022:DeepDPM}, 
a choice that is not justified in general. 

%To overcome  the above limitations, neural amortized models have been proposed recently that yield (approximate) 
To address these limitations, recent advancements have introduced neural amortized models that can generate (approximate)
i.i.d.~samples from $p(c_{1:N}|x_{1:N})$, 
without requiring an explicit generative model $p(x_i|\mu_{c_i})$. 
The Neural Clustering Process~(NCP)~\citep{pakman2020},
is a model for~$p(c_{1:N}|x_{1:N})$
that approximates the marginal posteriors \begin{align}
p(c_{1:n}| x_{1:N}  )=
\!\!\! \sum_{c_{(n+1):N}} 
\!\!\! 
p(c_{1:N}| x_{1:N} ) \,,
 %\quad n=1 \ldots N\!-\!1
 \label{eq:posterior_marginals}
\end{align} 
for $n=1 \ldots N-1$, 
using an energy-based model,
\begin{equation}
p_{\theta}(c_{1:n}| x_{1:N}  ) = 
  \frac{e^{-E_n[c_{1:n}, x_{1:N}]}}{Z_n[x_{1:N}]} \,,
\label{eq:NCP_marginals}
\end{equation}
where $Z_n[x_{1:N}]$ is a normalization constant
and $\theta$ are the parameters of the energy 
function~$E_n$. Note that under the  generative model~(\autoref{eq:gen1}),
the marginal posteriors~(\autoref{eq:posterior_marginals})
depend on all the  data points   $x_{1:N}$, 
not just on $x_{1:n}$.
Using the approximate marginals (\autoref{eq:NCP_marginals}),
posterior samples are obtained by fixing a data order and  sequentially sampling from  each factor in the approximate expansion:
\eqan 
p_{\theta}(\bc |\bx) 
\simeq   p(c_1| \bx ) p_{\theta}(c_2|c_1,\bx ) \ldots p_{\theta}(c_N|c_{1:N-1}, \bx)\,,
\label{joint}
\enan
given by:  
\begin{align}
p(c_1=1| \bx ) &= 1
\label{eq:first_assgn}
\\
p_{\theta}(c_n|c_{1:n-1}, \bx) 
&= \frac{  p_{\theta}(c_{1:n} |\bx)}
{  \sum_{c_n'=1}^{K+1}  p_{\theta}(c_1 \ldots c_n'| \bx)},
\label{eq:conditional}
\\
&= \frac{  e^{-E_n[c_{1:n}, x_{1:N}]} }
{  \sum_{c_n'=1}^{K+1}  
e^{-E_n[c_1 \ldots c_n', x_{1:N}]}}.
\label{eq:conditional2}
\end{align}
Note that the first assignment (\autoref{eq:first_assgn}) is deterministic. The NCP setup assumes $K$ unique values in $c_{1:n-1}$,
so~$c_n$ can take $K+1$ values, i.e., $x_n$ can join any existing cluster or form its own new cluster. 
Interestingly, the normalization constant from \autoref{eq:NCP_marginals} cancels out in the conditionals (Eq. \ref{eq:conditional}-\ref{eq:conditional2}). This allowed the authors of~\cite{pakman2020} to train the model via maximum likelihood,  
\begin{align}
    \theta = \argmax_{\theta} \mathbb{E}_{ p_{\text{data}}(\bc, \bx)} [\log p_{\theta}(\bc |\bx) ]    \,,
\end{align}
without contrastive divergence~\citep{hinton2002training}. 
A major limitation of the NCP model, however, is that 
although \autoref{eq:conditional} assumes 
the validity of \autoref{eq:marginal_consistency1}, the latter is not enforced 
in the architecture or the loss. 

\paragraph{Architecture of the energy function.} The NCP energy function $E_n[c_{1:n}, x_{1:N}]$ in \autoref{eq:NCP_marginals}
is the scalar output of 
a network that respects the following symmetries of the marginal posteriors $p(c_{1:n}| x_{1:N})$ in \autoref{eq:posterior_marginals}: 
% The general problem of constructing such invariant encodings was discussed recently in \citep{deep_sets}; to adapt this approach to our context, we consider three distinct permutation symmetries:
\newlist{myitemize}{itemize}{3}
\setlist[myitemize,1]{label=\textbullet,leftmargin=0.2in}
\begin{myitemize}

	\item Permutations {\bf within a cluster}	
 are preserved by encoding the data points in each cluster as: 
	\eqan
	H_k= \sum_{i : c_{i}=k} h(x_i)  %\qquad k = 1\ldots K\,,
	\quad h:\mathbb{R}^{d_x} \rightarrow \mathbb{R}^{d_h}
	\label{Hk}
	\enan 
	% which is clearly invariant under permutations of $x_i$'s in the same cluster. 

 \item Permutations {\bf  between clusters}
	%	\\
	% (\ref{eq:posterior_marginals}) is invariant under permutations of the cluster labels.
 are preserved, using the cluster invariants~$H_k$,  by 
 %in terms of the , 
	\eqan 
	G_{c_{1:n}}= \sum_{k =1}^K g(H_k) ,
	\quad 
	g:\mathbb{R}^{d_h} \rightarrow \mathbb{R}^{d_g}.
	\label{G_def}
	\enan 			

	\item Permutations of the {\bf  unassigned data points} are preserved  by 
	\eqan 
	U_{n+1} = \sum_{i=n+1}^{N}  u(x_i) \,,
		\qquad 
	u:\mathbb{R}^{d_x} \rightarrow \mathbb{R}^{d_u}.
	\label{Q}
	\enan 		
	\end{myitemize} 

The functions $h,g,u$ are neural networks.  
$G$ and~$U$ provide distributed, 
symmetry-invariant representations of the assigned and unassigned data points, respectively, for any~$N$ and $K$.  
Encodings of this form 
yield arbitrarily accurate approximations 
of (partially) symmetric 
functions~\citep{deep_sets,gui2021pine}.
Using the $G,U$ encodings, 
the NCP model represents the energy functions in \autoref{eq:NCP_marginals} as 
\eqan 
E_n[c_{1:n}, x_{1:N}] = f(G_{c_{1:n}},U_{n+1}) \,,
\label{eq:unnormalized_marginal}
\enan 
where $	f:\mathbb{R}^{d_g + d_u} \rightarrow \mathbb{R}$ is a neural network.

\subsection{Generative Flow Networks} 
\label{sec:gflonet_review}
\begin{figure*}[t!]
	\begin{center}
		\fbox{		
\includegraphics[width=.77\textwidth]{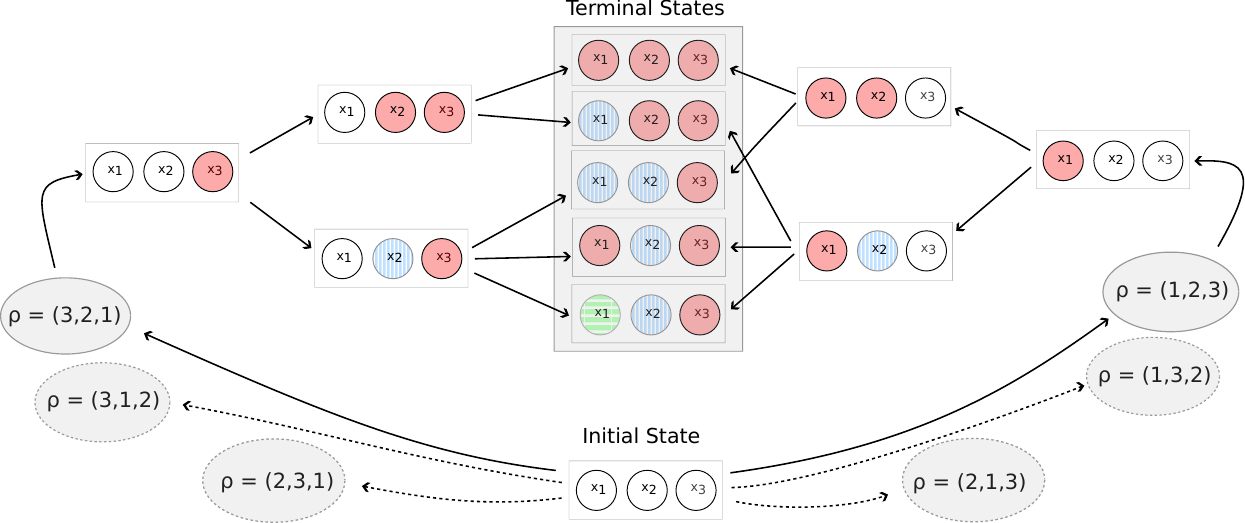}}
	\end{center}
	\caption{{\bf Clustering via sequential decisions}. 
 Directed Acyclical Graph (DAG) of sequential
 assignment of cluster labels  for a dataset of size $N=3$. The first action samples uniformly an order $\rho$ for the data points.  Each terminal state, corresponding to a fully clustered dataset, receives $N!$ incoming edges, corresponding to all possible orders of the data. 
	} 
	\label{fig:dag}	
\end{figure*}

GFlowNets \citep{bengio2021flow,bengio2023gflownet}
are a family of  models that amortize the 
cost of sampling over complex discrete objects. 
Assume we are given 
a directed acyclic graph (DAG) $(\mathcal{S}, \mathcal{A})$
where $\mathcal{S}$ is a finite set of vertices {\it (states)} and  $\mathcal{A} \subset \mathcal{S} \times \mathcal{S}$ is a
set of directed edges {\it (actions)}, and there is a unique initial state $s_0$ from which every other state is reachable. 

The composite objects of interest are represented by 
{\it terminal states} $\mathcal{Z} \subseteq \mathcal{S}$ without outgoing edges. Such objects can be constructed by following a trajectory 
$\tau = (s_0 \rightarrow s_1 \rightarrow \ldots
\rightarrow z)$, where $z \in \mathcal{Z}$. 
Denoting by $\mathcal{T}$ the set of trajectories, 
\cite{bengio2023gflownet} define a {\it trajectory flow} $F:\mathcal{T} \rightarrow \mathbb{R}^+$ as an unnormalized probability mass associated 
to trajectory~$\tau$. 
The {\it edge flow} is defined as  
$ F(s \rightarrow s') = \sum_{\tau: s \rightarrow s' \in \tau } F(\tau)$
and the {\it state flow} as 
$ F(s) = \sum_{\tau: s \in \tau } F(\tau)$.

Markovian GFlowNets treat the generation of a sample 
in $\mathcal{Z}$ 
as a Markov sequential decision problem. 
Samples from $\mathcal{Z}$ 
are obtained by starting from $s_0$ and 
sampling successive actions using a {\it forward policy}, 
which represents a distribution over the children of $s \in \mathcal{S} \backslash  \mathcal{Z}$ given~by 
\begin{align}
P_F(s'|s) = \frac{F(s \rightarrow s')}
{\sum_{s''} F(s \rightarrow s'') } \,,
\label{eq:forward}
\end{align}
until a terminal state is reached.  
% A {\it consistent flow} satisfies $\forall s \in \mathcal{S}$ the flow matching equations
% \begin{align}
%     \sum_{s' \in \text{ Parent}(s) } \!\!\!\!\!\! F( s' \rightarrow s)  = 
%     \!\!\!\!\!\!
%     \sum_{s'' \in \text{ Children}(s) } \!\!\!\!\!\! F(s \rightarrow s'').
% \label{eq:flow_match}
% \end{align}
% The goal of training a GFlowNet is to find a policy such that 
% the probability of reaching a terminal state $z \in \mathcal{Z}$ is proportional to a specified or learned {\it reward} function $R(z):\mathcal{Z} \rightarrow \mathbb{R}^+$, i.e., 
% \begin{align}
%     R(z) \varpropto  \sum_{z \in \tau } F(\tau) \,.
%     \label{eq:R_GFN}
% \end{align}
% As shown in~\cite{bengio2021flow}, this occurs 
% when using the policy of \autoref{eq:forward} in a consistent flow whose terminal states satisfy $F(z)\varpropto R(z)$. Given a parameterized edge flow, the Flow Matching loss~\citep{bengio2021flow} imposes  
% equation \autoref{eq:flow_match} in log space. 
% Recent  works also use other losses such as the Detailed Balance~\citep{bengio2023gflownet} or Trajectory Balance~\citep{malkin2022trajectory}, which exploit backward policies  over the parents of a state $s$. But in our case (see below) non-terminal states have a single parent. 
The goal of training a GFlowNet is to find a policy such that 
the probability of reaching a terminal state $z \in \mathcal{Z}$ is proportional to a specified or learned {\it reward} function $R(z) \in \mathbb{R}^+$, i.e., 
\begin{align}
    R(z) \varpropto  \sum_{\tau: z \in \tau } F(\tau) \,.
    \label{eq:R_GFN}
\end{align}
As shown in~\cite{bengio2021flow}, this occurs 
when using a policy of the form \autoref{eq:forward} obtained from edge flows 
satisfying the flow matching equations
\begin{align}
    \sum_{s' \in \text{ Parent}(s) } \!\!\!\!\!\! F( s' \rightarrow s)  = 
    \!\!\!\!\!\!
    \sum_{s'' \in \text{ Children}(s) } \!\!\!\!\!\! F(s \rightarrow s'')\,,
\label{eq:flow_match}
\end{align}
and whose terminal states satisfy $F(z)\varpropto R(z)$. Given a parameterized edge flow, the Flow Matching loss~\citep{bengio2021flow} imposes  
equation \autoref{eq:flow_match} in log space. 
Recent  works use other losses such as Detailed Balance~\citep{bengio2023gflownet}, Trajectory Balance~\citep{malkin2022trajectory}
or SubTB($\lambda$)~\citep{madan2023learning}, 
which exploit backward policies  over the parents of a state $s$. 
% But in our case (see below) non-terminal states have a single parent. 

% A GFlowNet is structurally equivalent to a Markov Reward Process~\citep{ronald1971dynamic}, i.e., 
% a Markov Decision Process with deterministic dynamics, and has close connections with entropy-regularized reinforcement learning~\citep{tiapkin2024generative}. 

\section{GFLOWNET CLUSTERING}
Given $N$ data points, we formalize sampling from the clustering posterior using the GFlowNet framework as follows: the initial action uniformly samples a data order, $\rho$, and assigns the first data point to the first cluster.
Then, a sequence of $N-1$ sampled actions 
allow each successive data point to join an existing cluster or create a new one. A terminal state $z=s_N$ represents a fully clustered dataset. As illustrated in~\Cref{fig:dag}, the initial state~$s_0$ edges out into~$N!$ branches, all of which  meet again in each of the terminal states. 

 The highly symmetric nature of our setting implies 
an additional property not present in general GFlowNets. 
For each terminal state $z$,
there are $N!$ trajectories~$\tau$
which differ by the data order. 
However, conditional exchangeability (\autoref{eq:exchange}) implies that all these trajectories are of equal probability.
Thus, 
 \autoref{eq:R_GFN} becomes:
\begin{align}
R(z) \varpropto F(\tau) \qquad \forall z \forall \tau: z  \in \tau    \,.
\label{eq:R_F}
\end{align}
The reward itself, $R(z) \varpropto p(\bc|\bx)$, is learned from the data. 
This setting was studied
in~\cite{zhang2022generative}, where an energy-based model
for the reward was learned together with the sampling policy. 
% We follow a similar approach here, with a major difference;
% instead of training separate networks for the policy and the reward, we use a common parametrization for both.
We take a similar approach with one key difference: rather than training separate networks for the policy and the reward, we use a shared parametrization for both.
% Such a possibility, originally suggested in~\cite{bengio2023gflownet}, seems not to have been made concrete before. 
This idea, originally suggested in~\cite{bengio2023gflownet}, does not appear to have been concretely developed before.
%To achieve this, we depart from the typical 
%approach in GFlowNet models
%to parameterize the forward policy $P_F(-|s)$
%as a neural network that takes a representation of $s$ as input and produces the logits of a distribution over its children as outputs. 
%
To achieve this, we deviate from the standard GFlowNet approach, where the forward policy $P_F(-|s)$ is parameterized as a neural network
that takes 
% a representation of 
$s$ as input,
and outputs 
%the logits of a distribution over its children.
logits for its children's distribution.
Instead, we exploit the NCP energy function~(\autoref{eq:unnormalized_marginal}) 
and use the same object to parameterize
state flows, edge flows and the reward. 
Concretely, given $N$ data points, let us define:
\begin{align}
\hat{E}[c_{1:n}] 
= 
\begin{cases}
0 & n=1,
\\
E[c_{1:n}] 
- 
\underset{c_n'}{\min}
{E}[c_{1:n-1},c_{n}']
& {}_{2 \leq n < N},  
\\
E[c_{1:N}] & n=N,  
\end{cases}
\nonumber
\end{align}
where $E[c_{1:n}] = f(G_{c_{1:n}},U_{n+1})$ is the energy function given the labels and the data, as described in~\autoref{eq:unnormalized_marginal}. 
We omit indicating the data $x_{1:N}$ in the energy arguments to simplify the notation. 
We define a state $s_{1 \leq n < N}$ as  $(\rho, c_{1:n})$, containing both the data order $\rho$ and the $n$ initial label assignments. 
The initial transition 
$s_0 \rightarrow s_1$ is special 
because it uniformly samples a permutation~$\rho$ for the data order. 
We parameterize the initial edge flow as $F[ s_0 \rightarrow s_1 = (\rho, c_{1} =1)]
= 1$, yielding a uniform forward transition~(\autoref{eq:forward})
$P_F[ c_1=1, \rho |s_0 ] = \frac{1}{N!}$.
% \begin{align}
%   P_F[ c_1=1, \rho |s_0 ] = \frac{1}{N!} \,.
% \end{align}
For non-initial states, we model both edge and state flows using the {\it same} function (for $2 \leq n \leq N$): 
\begin{align}
F[ c_{1:n-1} \rightarrow
c_{1:n}, \rho] & = F[ c_{1:n}, \rho] = e^{-\hat{E}[c_{1:n}]}  \,.
% \\
% F[ c_{1:n}, \rho] & = e^{-\hat{E}[c_{1:n}]}  
% \quad n=2 \ldots N.
\label{eq:FFE}
\end{align}
This yields the forward transition (\autoref{eq:forward})
\begin{align}
   P_F[ c_n | c_{1:n-1}, \rho] =  
   \frac{e^{-\hat{E}[c_{1:n}]}}
   {\sum_{c_n'}  
   e^{-\hat{E}[c_{1:n-1},c_n']} }\,,
\label{eq:forward_clustering}
\end{align}
for $2 \leq n \leq N$, and the order-independent reward
\begin{align}
% p(\bc|\bx) \varpropto
R[c_{1:N}] & \varpropto F[c_{1:N}] = e^{-{E}[c_{1:N}]} \,. 
\label{eq:reward}
\end{align}

\subsection{Objective function}\label{seq:obj_func} 
To train the model, we consider a loss function over the network parameters $\btheta$ that consists of three terms:

{\bf Marginal Consistency loss.} 
Considering \autoref{eq:FFE} and the fact that 
non-terminal states have a single parent, 
the flow matching equations (\autoref{eq:flow_match}) become in our case
\begin{align}
F [c_{1:n-1}, \rho] = \sum_{c_{n}} F [c_{1:n}, \rho],
\label{eq:flow_match_for_us}
\end{align}
and our goal is to minimize the discrepancy between the two terms.
By substituting $F[ c_{1:n}, \rho] = e^{-\hat{E}[c_{1:n}]}$ (as shown in~\autoref{eq:FFE}), and transitioning to log space, we aim to minimize the following loss function:
%We approximate this equation in log space along a trajectory~$\bc$ via the 
%loss 
%($\btheta$ are the parameters of $\hat{E}$)
% \begin{align}
% \mathcal{L}_{\btheta}^{mc}(\bc,\bx)& =\Sigma_{n=2}^{N} \mathcal{L}_{\btheta}^{mc}(c_{1:n-1})\,,
% \label{eqn:mc_loss}
% \end{align}
% where
% \begin{align}
% \mathcal{L}_{\btheta}^{mc}(c_{1:n-1}) = 
% \left( \hat{E}[c_{1:n-1}]  +\log \sum_{c_{n}}
% e^{-\hat{E}[c_{1:n}]}
% \right)^2.
% \nonumber
% \end{align}
\begin{align}
\mathcal{L}_{\btheta}^{\text{mc}}(\bc,\bx)& \!=\!\sum_{n=2}^{N} 
\left( \hat{E}[c_{1:n-1}]  
+\log \Sigma_{c_{n}}
e^{-\hat{E}[c_{1:n}]}
\right)^2\!\!\!.
\label{eqn:mc_loss}
\end{align}
This is the Flow Matching loss~\citep{bengio2021flow}, which we dub {\it Marginal Consistency} loss.
In our model it is preferred over the other 
losses mentioned 
in~\autoref{sec:gflonet_review} (Detailed Balance, Trajectory Balance and SubTB($\lambda$))
because \autoref{eqn:mc_loss} directly enforces the consistency of marginal distributions, which is central to the order invariance 
proof in~\autoref{subsec:order_invariance}. The other losses would only enforce this condition asymptotically, achieving equivalence only in the limit of large model capacity and training data.

%%%%%% Algorithm (training) %%%%%%%%
\begin{algorithm}[t!]
\caption{GFNCP training framework}
\label{alg:training}
\textbf{Input:} Training distribution  $p_{\text{data}}(\bc,\bx)$ over data $\bx =\{x_i\}_{i=1}^N$ and assignments $\bc =\{c_i\}_{i=1}^N$ (using instance discrimination, \autoref{seq:self_sup}), hyperparameter $\beta \in [0,1]$. 
\begin{algorithmic}[1]
    \State Initialize $u, h, g,f$ networks with parameters $\btheta$
    \Repeat
        \State $l \sim \text{Bernoulli}(\beta)$
        \If{$l = 1$}   \algorithmiccomment{Data policy}
             \State Sample $(\bc,\bx) \sim p_{\text{data}}(\bc,\bx)$
            \State \parbox[t]{180pt}{Update $\btheta$ with gradient 
            \\
            $\nabla_{\btheta}[\mathcal{L}^{\text{mc}}_{\btheta}(\bc,\bx) + \delta 
            \mathcal{L}^{\text{reg}}_{\btheta}(\bc,\bx) + \lambda \mathcal{L}^{\text{cd}}_{\btheta}(\bc,\bx)]$ 
            \strut}
        \Else  \algorithmiccomment{Space exploration}
             \State Sample $\bx \sim p_{\text{data}}(\bx)$
            \State \parbox[t]{180pt}{Uniformly sample $\bc^U$.  \strut}
            \State \parbox[t]{180pt}{Update $\btheta$ with gradient 
            \\
            $\nabla_{\btheta}[\mathcal{L}_{\btheta}^{\text{mc}}(\bc^U,\bx) + \delta  
            \mathcal{L}_{\btheta}^{\text{reg}}(\bc^U,\bx)]$ }
        \EndIf
    \Until{convergence criterion}
\end{algorithmic}
\end{algorithm}
%%%%%% END Algorithm (training) %%%%%%%%

{\bf Reward loss.}
The normalized reward is: 
\begin{align}
    p_{\btheta}(c_{1:N}|\bx) = \frac{e^{-E[c_{1:N}]}}{Z},
\end{align}
where $Z$ is a normalization constant.  
To learn this function we minimize the negative log-likelihood,
\begin{equation}
\mathcal{L}_{\btheta}^{\text{cd}}(\bc,\bx)=-  \  \log p_{\btheta}(c_{1:N}|\bx) \,,
\label{eqn:cd_loss}
\end{equation}
whose contrastive divergence 
gradient is: 
\begin{align}
    \nabla_{\theta} \mathcal{L}_{\btheta}^{\text{cd}}(\bc,\bx) 
    = \nabla_{\theta} E[c_{1:N}]  - 
    \mathbb{E}_{p_{\btheta}(\tilde{c}_{1:N}|\bx) }
    E[\tilde{c}_{1:N}] \,.
\label{eq:cd_gradient}
\end{align}
In the second term we approximate the expectation using a sample $\tilde{c}_{1:N}$ generated by the learned policy. 

{\bf Regularization.} To regularize the value of the energy function we add the following term:
\begin{equation}
\mathcal{L}_{\btheta}^{\text{reg}}(\bc,\bx)
=(E[c_{1:N}])^2. 
\label{eqn:reg_equation}
\end{equation}
{\bf Training data and policy.}   
We create a training distribution $p_{\text{data}}(\bc,\bx)$ as described below in~\autoref{seq:self_sup}. Since each training data point $(\bc,\bx)$ contains a full trajectory, we found it convenient to optimize the losses  {\it off-policy} (except for $\tilde{\bc}$ in \autoref{eq:cd_gradient}). 
The full objective  is:
\begin{align}
\mathcal{L}_{\btheta} 
& =
\mathbb{E}_{\pi(\bc|\bx)p_{\text{data}}(\bx) }
[\mathcal{L}_{\btheta}^{\text{mc}}(\bc,\bx)
+ \delta \mathcal{L}_{\btheta}^{\text{reg}}
(\bc,\bx)
] 
\label{eqn:combined_loss}
\\ & + 
\nonumber
\lambda
\mathbb{E}_{p_{\text{data}}(\bc,\bx) }
 [\mathcal{L}_{\btheta}^{\text{cd}}(\bc,\bx)], 
\end{align}
where $\delta, \lambda$ are hyperparameters and the policy $\pi(\bc|\bx)$ is a 
mixture of $p_{\text{data}}(\bc|\bx)$ and 
a uniform random sample of $\bc$, to assure full support 
for the flow matching equations~\autoref{eq:flow_match_for_us}. See~\autoref{alg:training}.

% \RED{
% Suggestion (Irit):

% Marginal-consistency loss:
% \begin{align}
% \mathcal{L}_{\btheta}^{mc}(\bc,\bx)& =\Sigma_{n=1}^{N-1} (log\ \tilde{p_{\btheta}}(c_{1:n-1}|\bx)\\
% & - log \Sigma_{c_n}\tilde{p_{\btheta}}(c_{1:n-1},c_n|\bx))^2 
% \end{align}

% Contrastive-divergence loss:
% \begin{equation}
% \mathcal{L}_{\btheta}^{cd}(\bc,\bx)=- \ \lambda (log\ p_{\btheta}(c_{1:N-1}|\bx)) 
% \end{equation}

% Regularization:
% \begin{equation}
% \mathcal{L}_{\btheta}^{reg}(\bc,\bx)=\delta (log\ \tilde{p_{\btheta}}(c_{1:N-1}|\bx))
% \end{equation}

% Full objective function:

% $$
% \mathcal{L}_{\btheta}(\bc,\bx)=\mathcal{L}_{\btheta}^{mc} + \mathcal{L}_{\btheta}^{cd} + \mathcal{L}_{\btheta}^{reg}
% $$

% }

% \subsection{Data order in the state definition}
% Similar to~\cite{zhang2023let}, we fix a uniform backward policy $P_B(q_n|s_n) = \frac{1}{n}$. 

\subsection{Order invariance}
\label{subsec:order_invariance}
{\bf Lemma:} when \autoref{eqn:mc_loss} is zero for all trajectories, the probability of reaching a final state is independent of the selected data order $\rho$, and thus \autoref{eq:R_F} holds. 
\begin{proof}
The probability of any trajectory $\tau$ is 
\begin{align}
\label{eq:prob_trajectory}
P(\tau) = & P_F(c_1 ) P_F(c_2| c_1 ) 
 \ldots P_F(c_N|c_{1:N-1})
\\
 & = \frac{1}{N!} 
\frac{ \cancel{ F(c_{1:2}, \rho) }}
{ \cancel{ \sum\limits_{c_2'} F(c_1, c_2', \rho) }}
\cdots
\frac{ F(c_{1:N}) } 
{ \cancel{ \sum\limits_{c_N'}
F(c_{1:N-1},c_N', \rho)}}  
\nn 
\\
& \varpropto F(c_{1:N}) = e^{-E[c_{1:N}] },
\label{eq:result_cancellations}    
\end{align}
where in all the pairwise cancellations we used \autoref{eq:flow_match_for_us},
which holds when \autoref{eqn:mc_loss} is zero. 
The proof is completed by noting that 
$E[c_{1:N}]$ is invariant 
under identical simultaneous permutations of $x_{1:N}$
and $c_{1:N}$.
\end{proof}
% The proof is similar to that of Proposition 2 in  \citealt{bengio2021flow}, 

\subsection{Self-supervised learning} \label{seq:self_sup}   
While traditional amortized clustering models require ground-truth 
labels~\citep{pakman2018amortized,pakman2020, DAC}, we propose a self-supervised approach using instance discrimination~\citep{dosovitskiy2014discriminative,wu2018unsupervised,chen2020simple,he2020momentum}, 
which assigns unique labels to each data point based on data augmentations. 

\paragraph{Computational cost.}
Like NCP, our model has an inference cost of $O(N)$. Faster models, which sample cluster members in parallel with $O(K)$ cost \citep{pakman2020}, introduce latent continuous variables and do not yield sample probabilities,  as this requires expensive marginalizations.

\section{EXPERIMENTS}

\begin{figure*}[t!]
	\begin{center}
		\fbox{		
		\includegraphics[width=.9\textwidth]{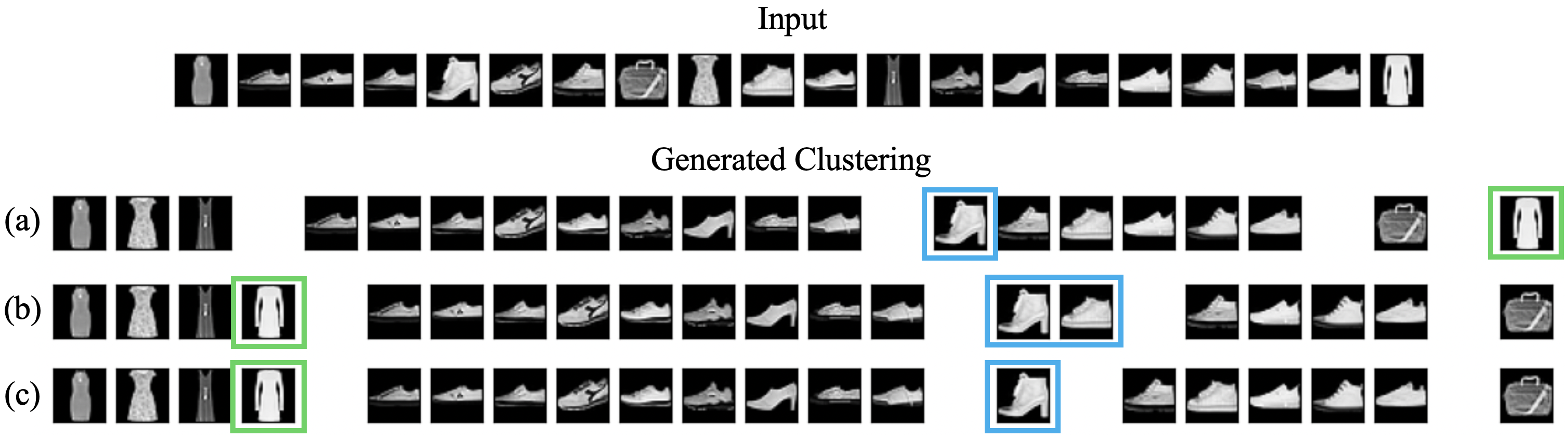}}
	\end{center}
	\caption{Top-three most-likely clusterings (\ie $p(a)\geq p(b)\geq p(c)$) generated by GFNCP, trained on the Fashion-MNIST dataset. We show that GFNCP's predictions are influenced by point interactions: 
    the model captures the data  uncertainty when assigning the dress (green) and the boot (blue), highlighting their distinctiveness within the context of the input set.} 
	\label{fig:fmnist_generated_clustering}	
\end{figure*}

% -------------------------------------------------
%                  General Comments
% -------------------------------------------------

We present an extensive evaluation of GFNCP on several datasets,
covering synthetic and toy data in \autoref{subsec:toy_data}, and real-world data in \autoref{subsec:real_data}.
We compare GFNCP with other notable existing approaches, including Set Transformer (ST)~\citep{lee2019set}, DAC~\citep{DAC} and NCP~\citep{pakman2020}. 
We demonstrate that GFNCP outperforms existing methods in clustering performance and maintains invariance to input data order, supported by both quantitative and qualitative evidence.
%
% In \autoref{subsec:toy_data} we analyze GFNCP's clustering performance and data-permutation invariance on synthetic and toy data. In \autoref{subsec:real_data} we continue evaluating GFNCP on real-data (embedded ImageNet~\citep{Russakovsky:IJCV:2015:imagenet})
% % using arbitrary number of clusters. 
% %
% We compare GFNCP with other notable existing approaches, including Set Transformer (ST)~\citep{lee2019set}, DAC~\citep{DAC} and NCP~\citep{pakman2020}. 
% %
% We demonstrate that GFNCP is invariant to the order of input data, supported by both quantitative and qualitative results. 
%
In all of our experiments we report the average values on three runs, selecting the best model based on the lowest loss value.

\subsection{Set-structured input generation}\label{subsec:input_set_generation}

Recall that our method can be trained in an unsupervised manner (\autoref{seq:self_sup}). 
In all our experiments, we generate the training-data distribution $p_{\text{data}}(\bc,\bx)$ by following~\autoref{eq:gen1} and utilizing a Chinese Restaurant Process (CRP)~\citep{pitman_csp} for clustering mixtures, where each cluster comprises a sampled data point.
Specifically, we adhere to the following steps: (1) the data size $N$ is either fixed or sampled,  and 
a clustering mixture $\bc$ is drawn using the CRP; (2) for self-supervision, instance discrimination is applied: 
data for each cluster contains a random data point from the dataset plus its augmentations to match the cluster size.  
For the test data, we follow the same procedure, except that in step (2), we use the original data points instead of augmentations.
See~\autoref{sec:data_generation} for details. 
% Further details on data generation and CRP parameters are available in~\autoref{sec:data_generation}.

% -------------------------------------------------
%                Invariance Metrics
% -------------------------------------------------

\subsection{Data-order invariance metrics}\label{subsec:inv_metrics}

% To evaluate the gap between our model and NCP, we introduce two quantitative metrics for evaluating data-permutation invariance.
%To assess the discrepancy between our model and NCP, we define two quantitative indicators to assess data-permutation invariance.
% \indent \textbf{Marginal Consistency (MC).} This metric measures the marginal-consistency loss (\autoref{eqn:mc_loss}) based on the marginal probabilities at test time.
% \indent \textbf{Standard Deviation over Permutation Probabilities (SDPP).} For this metric, we conduct an experiment where we generate the model's output probabilities given 500 different permutations of the same dataset. We then compute the standard deviation (SD) of the resulting probabilities and normalize it by dividing by the mean (M) probability, to eliminate scaling bias in method comparisons.
%
% The SDPP metric is defined as follows:
%         \begin{equation}
%             \text{SDPP} = \mathbb{E}_{p(N)p(\bx)} \left[ \frac{SD ( \{ p(\bc_{\pi} | \bx_{\pi}) \}_{\pi \in \rho})}{M ( \{ p(\bc_{\pi} | \bx_{\pi}) \}_{\pi \in \rho})} \right]
%             \label{eq:inv_metric}
%         \end{equation}
% where $\rho$ is a set of 500 random permutations. 
% Lower values of the MC and SDPP metrics indicate greater consistency of the model. Note that ST and DAC are excluded from this comparison since they do not produce marginal probabilities for clustering assignments.
To compare GFNCP with NCP, we introduce two metrics. (i)
\textbf{Marginal Consistency (MC):} the average of the marginal-consistency loss (\autoref{eqn:mc_loss}) 
over test pairs $(\bc,\bx) \sim p_{\text{data}}(\bc,\bx)$. 
%\indent 
(ii) \textbf{Standard Deviation over Permutation Probabilities (SDPP):} 
Given a pair $(\bc,\bx) \sim p_{\text{data}}(\bc,\bx)$, we evaluate the probabilities (\autoref{eq:prob_trajectory}) for 500 different permutations $(\bc_{\rho},\bx_{\rho})$; we then compute the standard deviation~(SD) of these probabilities and divide by their mean (M), to eliminate scaling bias.
The SDPP metric for $(\bc,\bx)$ is 
        \begin{equation}
            \text{SDPP}(\bc,\bx) =  
            \frac{SD ( \{p(\bc_{\rho} | \bx_{\rho})\}_{\rho \in S^N} ) }
            {M ( \{p(\bc_{\rho} | \bx_{\rho})\}_{\rho \in S^N} ) }
            \label{eq:inv_metric} \,.
        \end{equation}
% and the SDPP metric for the model is 
%         \begin{equation}
%           \text{SDPP} =\mathbb{E}_{p_{data}(\bc,\bx)}[\text{SDPP}(\bc,\bx)]  
%             \label{eq:inv_metric}
%         \end{equation}
Lower MC and SDPP metrics indicate greater consistency of the model. Note that ST and DAC are excluded from this comparison since they do not produce probabilities for clustering assignments.

% -------------------------------------------------
%                  Toy data
% -------------------------------------------------

\begin{figure*}[hbt!]  %[t!]
	\begin{center}	
		\includegraphics[width=.8\textwidth]{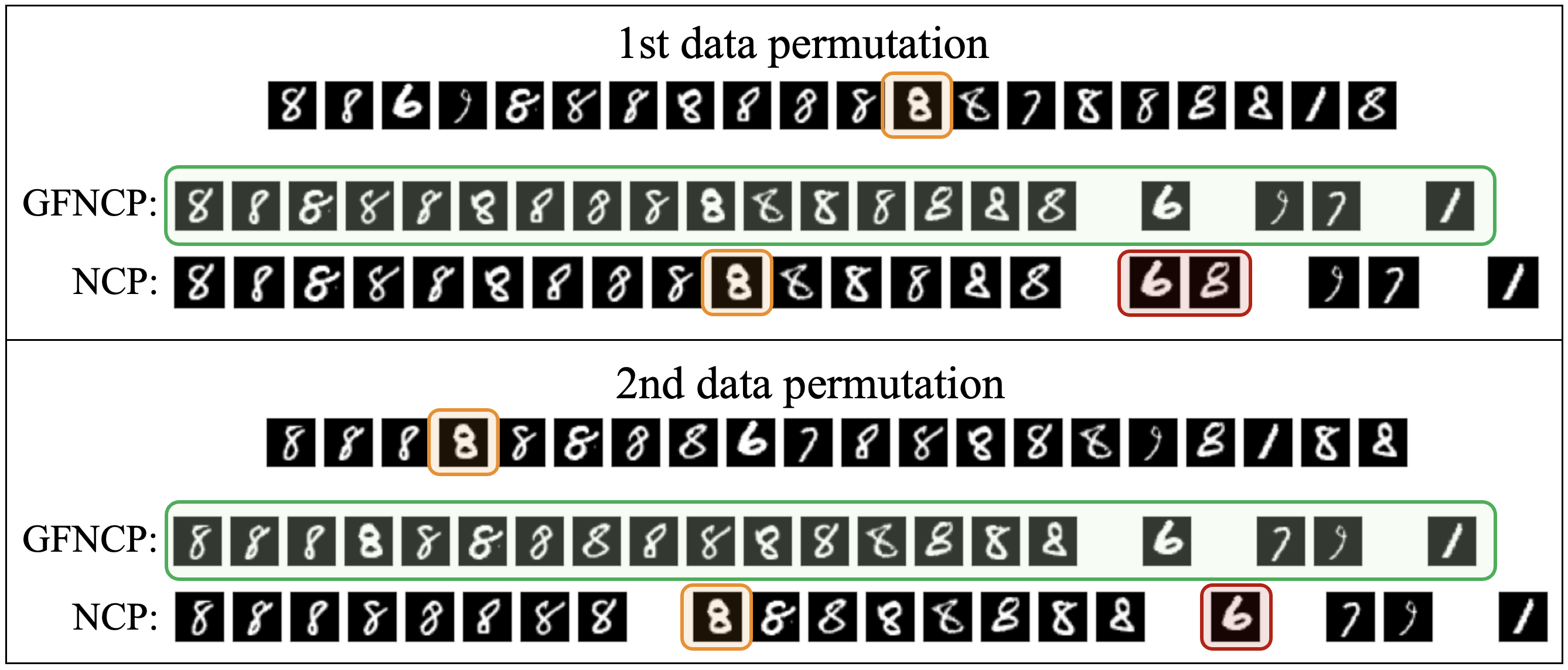}
	\end{center}
	\caption{Most-likely assignments generated by GFNCP and NCP, given two different data orders of the same input. GFNCP demonstrates consistent predictions (highlighted in green), whereas NCP’s results vary based on the order of the data (with an example shown in red). The digits marked in orange are discussed in the text.} 
	\label{fig:perms_GFNCP_vs_NCP_mnist}	
\end{figure*}

% \pm

\begin{table}[hbt!] %[ht]
    \centering
    %\footnotesize
    \setlength{\abovetopsep}{1.3ex}
    \caption{Clustering results on test sets sampled from MoG, MNIST and Fashion-MNIST (FMNIST), using {\bf a fixed number of six clusters.}}
    \resizebox{1\linewidth}{!}{ 
    \begin{tabular} {lccccc}
         \toprule
         \multicolumn{1}{l}{METHOD} & & ST & DAC & NCP & GFNCP \\
         \midrule
          \multirow{3}{*}{\parbox{1cm}{\textbf{MoG}}} & \small{NMI} & 0.91 \footnotesize{$\pm$ 0.03} & \textbf{0.96} \footnotesize{$\pm$ 0.01} & 0.95 \footnotesize{$\pm$ 0.01} & \textbf{0.96} \footnotesize{$\pm$ 0.00} \\
         & \small{ARI} & 0.91 \footnotesize{$\pm$ 0.04} & 0.97 \footnotesize{$\pm$ 0.01} & 0.95 \footnotesize{$\pm$ 0.01} & \textbf{0.98} \footnotesize{$\pm$ 0.01} \\
         & \small{MC $\downarrow$} & -- & -- & 10.3 \footnotesize{$\pm$ 7.7} & \textbf{3.7} \footnotesize{$\pm$ 2.5} \\
         \midrule
         \multirow{3}{*}{\parbox{1cm}{\textbf{MNIST}}} & \small{NMI} & 0.41 \footnotesize{$\pm$ 0.01} & 0.27 \footnotesize{$\pm$ 0.07} & 0.68 \footnotesize{$\pm$ 0.24} & \textbf{0.79} \footnotesize{$\pm$ 0.08} \\
         & \small{ARI} & 0.24 \footnotesize{$\pm$ 0.01} & 0.23 \footnotesize{$\pm$ 0.06} & 0.71 \footnotesize{$\pm$ 0.23} & \textbf{0.80} \footnotesize{$\pm$ 0.09} \\
         & \small{MC $\downarrow$} & -- & -- & 34.3 \footnotesize{$\pm$ 28.1} & \textbf{12.3} \footnotesize{$\pm$ 10.5} \\
         \midrule
        \multirow{3}{*}{\parbox{1cm}{\small{\textbf{FMNIST}}}} & \small{NMI} & 0.38 \footnotesize{$\pm$ 0.01} & 0.34 \footnotesize{$\pm$ 0.01} & 0.50 \footnotesize{$\pm$ 0.05} & \textbf{0.53} \footnotesize{$\pm$ 0.02} \\
         & \small{ARI} & 0.25 \footnotesize{$\pm$ 0.00} & 0.20 \footnotesize{$\pm$ 0.02} & 0.44 \footnotesize{$\pm$ 0.07} & \textbf{0.48} \footnotesize{$\pm$ 0.04} \\
         & \small{MC $\downarrow$} & -- & -- & 60.9 \footnotesize{$\pm$ 15.2} & \textbf{42.6} \footnotesize{$\pm$ 18.2} \\
         \bottomrule
    \end{tabular}
     }
    \label{tab:clustering_stats_toy_data_fixed_k}
\end{table}

% \pm

\begin{table}[hbt!] %[ht]
    \centering
    %\footnotesize
    \setlength{\abovetopsep}{1.3ex}
    \caption{Clustering results on test sets sampled from MoG, MNIST and Fashion-MNIST (FMNIST), with {\bf an arbitrary number of clusters.}}
    \resizebox{1\linewidth}{!}{ 
    \begin{tabular} {lccccc}
         \toprule
         \multicolumn{1}{l}{METHOD} & & DAC & NCP & GFNCP \\
          \midrule
         \multirow{3}{*}{\parbox{1cm}{\textbf{MoG}}} & \small{NMI} & \textbf{0.93} \footnotesize{$\pm$ 0.01} & 0.92 \footnotesize{$\pm$ 0.01} & \textbf{0.93} \footnotesize{$\pm$ 0.01} \\
         & \small{ARI}  & 0.90 \footnotesize{$\pm$ 0.01} & 0.89 \footnotesize{$\pm$ 0.04} & \textbf{0.91} \footnotesize{$\pm$ 0.02} \\
         & \small{MC $\downarrow$}  & -- & 50.8 \footnotesize{$\pm$ 7.1} & \textbf{41.1} \footnotesize{$\pm$ 19.8} \\
         \midrule
         \multirow{3}{*}{\parbox{1cm}{\textbf{MNIST}}} & \small{NMI} & 0.31 \footnotesize{$\pm$ 0.03} & 0.65 \footnotesize{$\pm$ 0.08} & \textbf{0.72} \footnotesize{$\pm$ 0.07} \\
         & \small{ARI}  & 0.24 \footnotesize{$\pm$ 0.03} & 0.64 \footnotesize{$\pm$ 0.06} & \textbf{0.74} \footnotesize{$\pm$ 0.05} \\
         & \small{MC $\downarrow$}  & -- & 39.3 \footnotesize{$\pm$ 26.9} & \textbf{16.4} \footnotesize{$\pm$ 9.3} \\
         \midrule
        \multirow{3}{*}{\parbox{1cm}{\small{\textbf{FMNIST}}}} & \small{NMI}  & 0.33 \footnotesize{$\pm$ 0.01} & 0.51 \footnotesize{$\pm$ 0.10} & \textbf{0.57} \footnotesize{$\pm$ 0.03} \\
         & \small{ARI}  & 0.18 \footnotesize{$\pm$ 0.02} & 0.48 \footnotesize{$\pm$ 0.11} & \textbf{0.51} \footnotesize{$\pm$ 0.07} \\
         & \small{MC $\downarrow$}  & -- & 62.2 \footnotesize{$\pm$ 18.4} & \textbf{41.1} \footnotesize{$\pm$ 13.1} \\
         \bottomrule
    \end{tabular}
     }
    \label{tab:clustering_stats_toy_data_unlim_k}
\end{table}
\subsection{Toy data}\label{subsec:toy_data}

We first demonstrate GFNCP's performance 
on a two-dimensional Gaussian-Mixture-Model (MoG), MNIST and Fashion-MNIST (FMNIST)~\citep{xiao:2017:FMNIST} datasets, using either fixed or arbitrary number of clusters $K$.
For data generation we follow the procedure in \autoref{subsec:input_set_generation}.
% During training, we create input sets by randomly sampling $N$ data points 
%from the training dataset, 
% based on the CRP mixture. 
We use $N\sim (100,1000)$ for set size, and a batch size of $64$. For the fixed-clusters experiment we condition the CRP prior on $K=6$.
As the ST method is mainly designed for embedded images, we utilize a data encoder (similar to the one used in our architecture) for these experiments, to enhance its performance.
We evaluate the models on $10,000$ input sets drawn from the test data for MNIST and FMNIST, and on $3,000$ newly-generated input sets for MoG, where each set is of size $N=300$. 
We report NMI and ARI metrics \citep{fahad:2014:NMI_ARI} for clustering evaluation.
For NCP and GFNCP, these metrics are calculated on assignments
generated via greedy decoding, where each assignment~$c_n$ 
in the sequence is selected according to the highest forward probability~\autoref{eq:forward_clustering}. 
See \autoref{sec:data_gen_and_experimental_setup} for more details on the experimental setup and data generation.
We also evaluate GFNCP using the average-score approach, where metrics are derived from multiple assignment samples (see \autoref{sec:average_score} for more details).

\autoref{tab:clustering_stats_toy_data_fixed_k} and \autoref{tab:clustering_stats_toy_data_unlim_k} present results for fixed and varying number of clusters, respectively.
GFNCP consistently outperforms existing methods on different datasets and settings. We observe that DAC's performance deteriorates when the data mixture is less balanced, a factor influenced by the CRP hyper-parameter.
Note that ST is excluded from \autoref{tab:clustering_stats_toy_data_unlim_k} because it is not designed for varying~$K$.
In \autoref{fig:fmnist_generated_clustering} we illustrate the top-three most probable clustering generated by GFNCP, highlighting its capability to leverage point interactions in assignment predictions.
In \autoref{fig:perms_GFNCP_vs_NCP_mnist} we showcase GFNCP's invariance to input-data order, \eg, when the digit `8' (highlighted in orange) appears earlier in the data sequence, NCP generates an additional cluster for the thicker `8' instances, whereas GFNCP produces consistent results.
The gap in order invariance is also shown in \autoref{fig:invariance_ecdf_mnist}, where we present the histogram and empirical CDF (ECDF) of the SDPP metric for both GFNCP and NCP on MNIST.
%
% Alongside the aforementioned experiments, we utilize a well-known set of tests known as Geweke's Tests~\citep{geweke2004getting} in the Appendix to demonstrate that GFNCP can effectively approximate the prior distribution $p(c_{1:N})$. 
% In the Appendix, we use a popular family of tests called Geweke's Tests~\citep{geweke2004getting}, and show that GFNCP can approximate the prior distribution $p(c_{1:N})$. 
%In addition, in \autoref{subsec:geweke_test} we show that GFNCP can approximate the prior distribution, by using a popular family of tests called Geweke's test.

% -------------------------------------------------
%                  Real-world data
% -------------------------------------------------

\begin{table}[hbt!]
    \centering
    %\footnotesize
    \setlength{\abovetopsep}{1.3ex}
    \caption{Comparison of GFNCP with existing methods on ImageNet-50/100/200 (IN50/IN100/IN200) using an arbitrary number of clusters.}
    \resizebox{1\linewidth}{!}{ 
    \begin{tabular} {lcccc}
         \toprule
         \multicolumn{1}{l}{METHOD} & & DAC & NCP & GFNCP \\
         \midrule
         \multirow{3}{*}{\parbox{1cm}{\textbf{IN50}}} & \small{NMI} & 0.53 \footnotesize{$\pm$ 0.01} & 0.58 \footnotesize{$\pm$ 0.08} & \textbf{0.62} \footnotesize{$\pm$ 0.08} \\
         & \small{ARI} & 0.27 \footnotesize{$\pm$ 0.07} & 0.43 \footnotesize{$\pm$ 0.12} & \textbf{0.48} \footnotesize{$\pm$ 0.06} \\
         & \small{MC $\downarrow$} & -- & 65.6 \footnotesize{$\pm$ 13.9} & \textbf{60.7} \footnotesize{$\pm$ 17.8} \\
         \midrule
        \multirow{3}{*}{\parbox{1cm}{\small{\textbf{IN100}}}} & \small{NMI}  & 0.53 \footnotesize{$\pm$ 0.02} & 0.50 \footnotesize{$\pm$ 0.10} & \textbf{0.60} \footnotesize{$\pm$ 0.05} \\
         & \small{ARI}  & 0.13 \footnotesize{$\pm$ 0.03} & 0.35 \footnotesize{$\pm$ 0.07} & \textbf{0.40} \footnotesize{$\pm$ 0.05} \\
         & \small{MC $\downarrow$}  & -- & 67.9 \footnotesize{$\pm$ 15.5} & \textbf{53.0} \footnotesize{$\pm$ 13.5} \\
         \midrule
         \multirow{3}{*}{\parbox{1cm}{\small{\textbf{IN200}}}} & \small{NMI}  & 0.57 \footnotesize{$\pm$ 0.03} & 0.46 \footnotesize{$\pm$ 0.09} & \textbf{0.58} \footnotesize{$\pm$ 0.04} \\
         & \small{ARI}  & 0.25 \footnotesize{$\pm$ 0.01} & 0.28 \footnotesize{$\pm$ 0.07} & \textbf{0.37} \footnotesize{$\pm$ 0.02} \\
         & \small{MC $\downarrow$}  & -- & 64.4 \footnotesize{$\pm$ 20.5} & \textbf{41.6} \footnotesize{$\pm$ 8.2} \\
         \bottomrule
    \end{tabular}
     }
    \label{tab:clustering_stats_IN_50_100_200}
\end{table}

\subsection{Real-world data}\label{subsec:real_data}
\begin{figure}[hbt!]
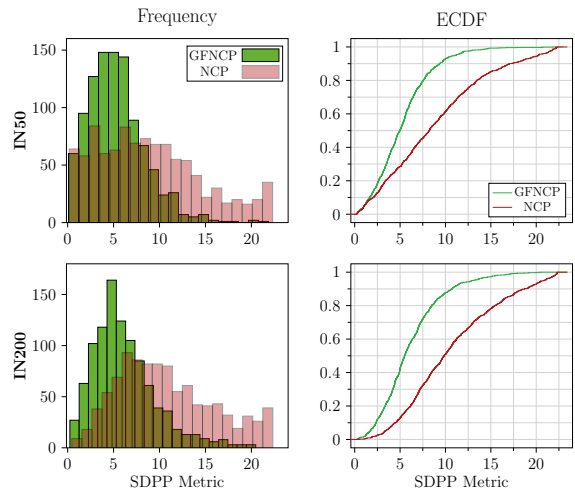
 %[H]
    % \renewcommand{\arraystretch}{0.04}
    % \centering
    \Large
    \begin{tabular}{p{3.6cm}p{3.6cm}}
    \scalebox{0.43}{
    \hspace{-0.8cm}
        % This file was created with tikzplotlib v0.10.1.
\begin{tikzpicture}

\definecolor{darkgray176}{RGB}{176,176,176}
\definecolor{lightblue}{RGB}{173,216,230}
% \definecolor{lightblue}{RGB}{173,216,230}
\definecolor{lightblue}{rgb}{0.6, 0.81, 0.93}
\definecolor{forestgreen4416044}{RGB}{44,160,44}
\definecolor{red_1}{RGB}{178,24,24}
\definecolor{green_1}{rgb}{0.4, 0.69, 0.2}

\begin{axis}[
tick align=outside,
tick pos=left,
x grid style={darkgray176},
title={\LARGE Frequency},
xmin=-0.13, xmax=24,
xtick style={color=black},
y grid style={darkgray176},
ylabel={\textbf{IN50}},
ymin=0, ymax=160,
ytick style={color=black},
y label style={at={(axis description cs:-0.16,.5)},anchor=south},
x label style={at={(axis description cs:.5,-0.12)}},
]

\node[text width=1cm] at (15,145.5) {\large GFNCP};
\node[text width=1cm] at (16.2,131.5) {\large NCP};
\draw[draw=black,fill=green_1] (axis cs:19,140) rectangle (axis cs:23,151);
\draw[draw=black,fill=red_1,opacity=0.4] (axis cs:19,125.5) rectangle (axis cs:23,136.5);
\draw[draw=black,fill=lightblue,fill opacity=0.0] (axis cs:13,123) rectangle (axis cs:23.4,153.5);

\draw[draw=black,fill=green_1] (axis cs:0.0935078626180651,0) rectangle (axis cs:1.18169446728288,60);
\draw[draw=black,fill=green_1] (axis cs:1.18169446728288,0) rectangle (axis cs:2.26988107194769,95);
\draw[draw=black,fill=green_1] (axis cs:2.26988107194769,0) rectangle (axis cs:3.3580676766125,127);
\draw[draw=black,fill=green_1] (axis cs:3.3580676766125,0) rectangle (axis cs:4.44625428127731,148);
\draw[draw=black,fill=green_1] (axis cs:4.44625428127731,0) rectangle (axis cs:5.53444088594212,148);
\draw[draw=black,fill=green_1] (axis cs:5.53444088594212,0) rectangle (axis cs:6.62262749060693,144);
\draw[draw=black,fill=green_1] (axis cs:6.62262749060693,0) rectangle (axis cs:7.71081409527174,89);
\draw[draw=black,fill=green_1] (axis cs:7.71081409527174,0) rectangle (axis cs:8.79900069993655,67);
\draw[draw=black,fill=green_1] (axis cs:8.79900069993655,0) rectangle (axis cs:9.88718730460136,46);
\draw[draw=black,fill=green_1] (axis cs:9.88718730460136,0) rectangle (axis cs:10.9753739092662,24);
\draw[draw=black,fill=green_1] (axis cs:10.9753739092662,0) rectangle (axis cs:12.063560513931,26);
\draw[draw=black,fill=green_1] (axis cs:12.063560513931,0) rectangle (axis cs:13.1517471185958,7);
\draw[draw=black,fill=green_1] (axis cs:13.1517471185958,0) rectangle (axis cs:14.2399337232606,5);
\draw[draw=black,fill=green_1] (axis cs:14.2399337232606,0) rectangle (axis cs:15.3281203279254,7);
\draw[draw=black,fill=green_1] (axis cs:15.3281203279254,0) rectangle (axis cs:16.4163069325902,2);
\draw[draw=black,fill=green_1] (axis cs:16.4163069325902,0) rectangle (axis cs:17.504493537255,1);
\draw[draw=black,fill=green_1] (axis cs:17.504493537255,0) rectangle (axis cs:18.5926801419198,1);
\draw[draw=black,fill=green_1] (axis cs:18.5926801419198,0) rectangle (axis cs:19.6808667465847,0);
\draw[draw=black,fill=green_1] (axis cs:19.6808667465847,0) rectangle (axis cs:20.7690533512495,2);
\draw[draw=black,fill=green_1] (axis cs:20.7690533512495,0) rectangle (axis cs:21.8572399559143,1);

\draw[draw=black,fill=red_1,opacity=0.4] (axis cs:0.173387967973786,0) rectangle (axis cs:1.28047465381509,64);
\draw[draw=black,fill=red_1,opacity=0.4] (axis cs:1.28047465381509,0) rectangle (axis cs:2.3875613396564,58);
\draw[draw=black,fill=red_1,opacity=0.4] (axis cs:2.3875613396564,0) rectangle (axis cs:3.49464802549771,84);
\draw[draw=black,fill=red_1,opacity=0.4] (axis cs:3.49464802549771,0) rectangle (axis cs:4.60173471133902,60);
\draw[draw=black,fill=red_1,opacity=0.4] (axis cs:4.60173471133902,0) rectangle (axis cs:5.70882139718032,63);
\draw[draw=black,fill=red_1,opacity=0.4] (axis cs:5.70882139718032,0) rectangle (axis cs:6.81590808302163,83);
\draw[draw=black,fill=red_1,opacity=0.4] (axis cs:6.81590808302163,0) rectangle (axis cs:7.92299476886294,69);
\draw[draw=black,fill=red_1,opacity=0.4] (axis cs:7.92299476886294,0) rectangle (axis cs:9.03008145470425,73);
\draw[draw=black,fill=red_1,opacity=0.4] (axis cs:9.03008145470424,0) rectangle (axis cs:10.1371681405456,68);
\draw[draw=black,fill=red_1,opacity=0.4] (axis cs:10.1371681405456,0) rectangle (axis cs:11.2442548263869,68);
\draw[draw=black,fill=red_1,opacity=0.4] (axis cs:11.2442548263869,0) rectangle (axis cs:12.3513415122282,55);
\draw[draw=black,fill=red_1,opacity=0.4] (axis cs:12.3513415122282,0) rectangle (axis cs:13.4584281980695,54);
\draw[draw=black,fill=red_1,opacity=0.4] (axis cs:13.4584281980695,0) rectangle (axis cs:14.5655148839108,41);
\draw[draw=black,fill=red_1,opacity=0.4] (axis cs:14.5655148839108,0) rectangle (axis cs:15.6726015697521,22);
\draw[draw=black,fill=red_1,opacity=0.4] (axis cs:15.6726015697521,0) rectangle (axis cs:16.7796882555934,30);
\draw[draw=black,fill=red_1,opacity=0.4] (axis cs:16.7796882555934,0) rectangle (axis cs:17.8867749414347,17);
\draw[draw=black,fill=red_1,opacity=0.4] (axis cs:17.8867749414347,0) rectangle (axis cs:18.993861627276,20);
\draw[draw=black,fill=red_1,opacity=0.4] (axis cs:18.993861627276,0) rectangle (axis cs:20.1009483131173,16);
\draw[draw=black,fill=red_1,opacity=0.4] (axis cs:20.1009483131173,0) rectangle (axis cs:21.2080349989586,20);
\draw[draw=black,fill=red_1,opacity=0.4] (axis cs:21.2080349989586,0) rectangle (axis cs:22.3151216847999,35);
\end{axis}

\end{tikzpicture}
    } & 
    \scalebox{0.43}{
        \hspace{-0.7cm}
        \input{./raw_inv_metric_ecdf_IN50}
    }\\
    \scalebox{0.43}{
    \hspace{-0.8cm}
        % This file was created with tikzplotlib v0.10.1.
\begin{tikzpicture}

\definecolor{darkgray176}{RGB}{176,176,176}
\definecolor{green_1}{RGB}{173,216,230}
% \definecolor{green_1}{RGB}{173,216,230}
\definecolor{green_1}{rgb}{0.6, 0.81, 0.93}
\definecolor{forestgreen4416044}{RGB}{44,160,44}
\definecolor{red_1}{RGB}{178,24,24}
\definecolor{green_1}{rgb}{0.4, 0.69, 0.2}

\begin{axis}[
tick align=outside,
tick pos=left,
x grid style={darkgray176},
xlabel={SDPP Metric},
xmin=-0.13, xmax=24,
xtick style={color=black},
y grid style={darkgray176},
ylabel={\textbf{IN200}},
ymin=0, ymax=180,
ytick style={color=black},
y label style={at={(axis description cs:-0.16,.5)},anchor=south},
x label style={at={(axis description cs:.5,-0.12)}},
]

\node[text width=1cm] at (6.4,290) {\large GFNCP};
\node[text width=1cm] at (6.9,265) {\large NCP};
\draw[draw=black,fill=green_1] (axis cs:8,280) rectangle (axis cs:9.5,300);
\draw[draw=black,fill=red_1,opacity=0.4] (axis cs:8,255) rectangle (axis cs:9.5,275);
\draw[draw=black,fill=green_1,fill opacity=0.0] (axis cs:5.5,249) rectangle (axis cs:9.7,306);

\draw[draw=black,fill=green_1] (axis cs:0.243007924568826,0) rectangle (axis cs:1.25457703660477,27);
\draw[draw=black,fill=green_1] (axis cs:1.25457703660477,0) rectangle (axis cs:2.26614614864072,63);
\draw[draw=black,fill=green_1] (axis cs:2.26614614864072,0) rectangle (axis cs:3.27771526067667,102);
\draw[draw=black,fill=green_1] (axis cs:3.27771526067667,0) rectangle (axis cs:4.28928437271262,118);
\draw[draw=black,fill=green_1] (axis cs:4.28928437271262,0) rectangle (axis cs:5.30085348474856,164);
\draw[draw=black,fill=green_1] (axis cs:5.30085348474856,0) rectangle (axis cs:6.31242259678451,124);
\draw[draw=black,fill=green_1] (axis cs:6.31242259678451,0) rectangle (axis cs:7.32399170882046,105);
\draw[draw=black,fill=green_1] (axis cs:7.32399170882046,0) rectangle (axis cs:8.33556082085641,85);
\draw[draw=black,fill=green_1] (axis cs:8.3355608208564,0) rectangle (axis cs:9.34712993289235,61);
\draw[draw=black,fill=green_1] (axis cs:9.34712993289235,0) rectangle (axis cs:10.3586990449283,39);
\draw[draw=black,fill=green_1] (axis cs:10.3586990449283,0) rectangle (axis cs:11.3702681569642,36);
\draw[draw=black,fill=green_1] (axis cs:11.3702681569642,0) rectangle (axis cs:12.3818372690002,18);
\draw[draw=black,fill=green_1] (axis cs:12.3818372690002,0) rectangle (axis cs:13.3934063810361,13);
\draw[draw=black,fill=green_1] (axis cs:13.3934063810361,0) rectangle (axis cs:14.4049754930721,13);
\draw[draw=black,fill=green_1] (axis cs:14.4049754930721,0) rectangle (axis cs:15.416544605108,9);
\draw[draw=black,fill=green_1] (axis cs:15.416544605108,0) rectangle (axis cs:16.428113717144,6);
\draw[draw=black,fill=green_1] (axis cs:16.428113717144,0) rectangle (axis cs:17.4396828291799,8);
\draw[draw=black,fill=green_1] (axis cs:17.4396828291799,0) rectangle (axis cs:18.4512519412159,3);
\draw[draw=black,fill=green_1] (axis cs:18.4512519412159,0) rectangle (axis cs:19.4628210532518,3);
\draw[draw=black,fill=green_1] (axis cs:19.4628210532518,0) rectangle (axis cs:20.4743901652878,3);

\draw[draw=black,fill=red_1,opacity=0.4] (axis cs:0.460160666903723,0) rectangle (axis cs:1.55391576978605,9);
\draw[draw=black,fill=red_1,opacity=0.4] (axis cs:1.55391576978605,0) rectangle (axis cs:2.64767087266837,18);
\draw[draw=black,fill=red_1,opacity=0.4] (axis cs:2.64767087266837,0) rectangle (axis cs:3.74142597555069,38);
\draw[draw=black,fill=red_1,opacity=0.4] (axis cs:3.74142597555069,0) rectangle (axis cs:4.83518107843301,54);
\draw[draw=black,fill=red_1,opacity=0.4] (axis cs:4.83518107843301,0) rectangle (axis cs:5.92893618131533,69);
\draw[draw=black,fill=red_1,opacity=0.4] (axis cs:5.92893618131533,0) rectangle (axis cs:7.02269128419765,93);
\draw[draw=black,fill=red_1,opacity=0.4] (axis cs:7.02269128419765,0) rectangle (axis cs:8.11644638707998,86);
\draw[draw=black,fill=red_1,opacity=0.4] (axis cs:8.11644638707998,0) rectangle (axis cs:9.2102014899623,82);
\draw[draw=black,fill=red_1,opacity=0.4] (axis cs:9.2102014899623,0) rectangle (axis cs:10.3039565928446,82);
\draw[draw=black,fill=red_1,opacity=0.4] (axis cs:10.3039565928446,0) rectangle (axis cs:11.3977116957269,80);
\draw[draw=black,fill=red_1,opacity=0.4] (axis cs:11.3977116957269,0) rectangle (axis cs:12.4914667986093,55);
\draw[draw=black,fill=red_1,opacity=0.4] (axis cs:12.4914667986093,0) rectangle (axis cs:13.5852219014916,61);
\draw[draw=black,fill=red_1,opacity=0.4] (axis cs:13.5852219014916,0) rectangle (axis cs:14.6789770043739,42);
\draw[draw=black,fill=red_1,opacity=0.4] (axis cs:14.6789770043739,0) rectangle (axis cs:15.7727321072562,41);
\draw[draw=black,fill=red_1,opacity=0.4] (axis cs:15.7727321072562,0) rectangle (axis cs:16.8664872101385,43);
\draw[draw=black,fill=red_1,opacity=0.4] (axis cs:16.8664872101385,0) rectangle (axis cs:17.9602423130209,32);
\draw[draw=black,fill=red_1,opacity=0.4] (axis cs:17.9602423130209,0) rectangle (axis cs:19.0539974159032,19);
\draw[draw=black,fill=red_1,opacity=0.4] (axis cs:19.0539974159032,0) rectangle (axis cs:20.1477525187855,31);
\draw[draw=black,fill=red_1,opacity=0.4] (axis cs:20.1477525187855,0) rectangle (axis cs:21.2415076216678,26);
\draw[draw=black,fill=red_1,opacity=0.4] (axis cs:21.2415076216678,0) rectangle (axis cs:22.3352627245502,39);
\end{axis}

\end{tikzpicture}
    } & 
    \scalebox{0.43}{
        \hspace{-0.7cm}
        \input{./raw_inv_metric_ecdf_IN200}
    }
    \end{tabular}
    \caption{Histogram and Empirical CDF (ECDF) of
the SDPP metric, computed on GFNCP’s and
NCP’s clustering results, trained on IN50/200 datasets.}    
    \label{fig:invariance_ecdf_IN50_IN200}
\end{figure}

We evaluate GFNCP's clustering performance on images sampled from ImageNet-50/100/200, which are subsets of 50/100/200 classes from ImageNet ~\citep{Russakovsky:IJCV:2015:imagenet}, using arbitrary $K$.
Since the models we consider assume that the cluster parameters,
$\mu_k$ (\autoref{eq:gen1}), have been integrated out, it is natural to explore 
their ability to generalize to unseen cluster classes.  
We thus sample training sets from half of the classes, and use the other half to form the test sets.
%
% To study the generalization of our method for unseen examples, training sets are sampled from half of the classes, while the other half is used to form the test sets, ensuring there is no overlap between the training and test classes.
%
Each image is represented by a 384-dim vector obtained from DINO~\citep{caron:DINO:2021}. 
Similar to the toy-data experiments, we follow the procedure in \autoref{subsec:input_set_generation} for data generation. %, with the same input-set sizes.
In \autoref{tab:clustering_stats_IN_50_100_200} we show a significant advantage to GFNCP. We report NMI, ARI and MC metrics, computed on the results over 2500 test sets.
In \autoref{fig:invariance_ecdf_IN50_IN200} we present the histogram and ECDF of the SDPP metric for GFNCP and NCP.
More information on the experimental setup can be found in \autoref{sec:data_gen_and_experimental_setup}.

%\subsection{Geweke’s Tests}\label{subsec:geweke_test}

In addition to the mentioned experiments, we evaluate GFNCP in an ``online mode". Unlike the primary mode, where the forward transition relies on the full dataset $x_{1:N}$, the model here sequentially predicts assignments for new data points based only on prior points and their cluster assignments, without access to future data. More details about this experiment are available in~\autoref{sec:online_mode}.

\section{CONCLUSION}\label{sec:conclusion}
In this paper, we presented GFNCP, a novel approach to amortized clustering that generates data assignments and their associated probabilities while accounting for point interactions and maintaining invariance to the order of the input data.
We showed that GFNCP outperforms existing methods in various tasks, in both clustering and consistency metrics.

\subsubsection*{Acknowledgments}
This work was supported in part by the Lynn and William Frankel Center at BGU CS, by Israel Science Foundation Personal Grant \#360/21, and by the Israeli Council for Higher Education (CHE) via the Data Science Research Center at BGU. 
A.P. was supported by the Israel Science Foundation (grant No. 1138/23).
I.C. was also funded in part by the Kreitman School of Advanced Graduate Studies, by BGU’s Hi-Tech Scholarship,
and by the Israel’s Ministry of Technology and Science Aloni Scholarship.

%\clearpage 
\bibliography{references} 

\clearpage
\section*{Checklist}

% % %%% BEGIN INSTRUCTIONS %%%
% The checklist follows the references. For each question, choose your answer from the three possible options: Yes, No, Not Applicable.  You are encouraged to include a justification to your answer, either by referencing the appropriate section of your paper or providing a brief inline description (1-2 sentences). 
% Please do not modify the questions.  Note that the Checklist section does not count towards the page limit. Not including the checklist in the first submission won't result in desk rejection, although in such case we will ask you to upload it during the author response period and include it in camera ready (if accepted).

% \textbf{In your paper, please delete this instructions block and only keep the Checklist section heading above along with the questions/answers below.}
% % %%% END INSTRUCTIONS %%%

 \begin{enumerate}

 \item For all models and algorithms presented, check if you include:
 \begin{enumerate}
   \item A clear description of the mathematical setting, assumptions, algorithm, and/or model. [Yes]
   \item An analysis of the properties and complexity (time, space, sample size) of any algorithm. [Yes]
   \item (Optional) Anonymized source code, with specification of all dependencies, including external libraries. [Not Applicable]
 \end{enumerate}

 \item For any theoretical claim, check if you include:
 \begin{enumerate}
   \item Statements of the full set of assumptions of all theoretical results. [Yes]
   \item Complete proofs of all theoretical results. [Yes]
   \item Clear explanations of any assumptions. [Yes]     
 \end{enumerate}

 \item For all figures and tables that present empirical results, check if you include:
 \begin{enumerate}
   \item The code, data, and instructions needed to reproduce the main experimental results (either in the supplemental material or as a URL). [Yes]
   \item All the training details (e.g., data splits, hyperparameters, how they were chosen). [Yes]
         \item A clear definition of the specific measure or statistics and error bars (e.g., with respect to the random seed after running experiments multiple times). [Yes]
         \item A description of the computing infrastructure used. (e.g., type of GPUs, internal cluster, or cloud provider). [Yes]
 \end{enumerate}

 \item If you are using existing assets (e.g., code, data, models) or curating/releasing new assets, check if you include:
 \begin{enumerate}
   \item Citations of the creator If your work uses existing assets. [Yes]
   \item The license information of the assets, if applicable. [Not Applicable]
   \item New assets either in the supplemental material or as a URL, if applicable. [Not Applicable]
   \item Information about consent from data providers/curators. [Not Applicable]
   \item Discussion of sensible content if applicable, e.g., personally identifiable information or offensive content. [Not Applicable]
 \end{enumerate}

 \item If you used crowdsourcing or conducted research with human subjects, check if you include:
 \begin{enumerate}
   \item The full text of instructions given to participants and screenshots. [Not Applicable]
   \item Descriptions of potential participant risks, with links to Institutional Review Board (IRB) approvals if applicable. [Not Applicable]
   \item The estimated hourly wage paid to participants and the total amount spent on participant compensation. [Not Applicable]
 \end{enumerate}

 \end{enumerate}

\clearpage
\appendix
\onecolumn

\section{DATA GENERATION AND EXPERIMENTAL SETUP}
\label{sec:data_gen_and_experimental_setup}

Here we provide technical details about the experiments mentioned in the paper, including input-data generation and training procedures. Recall that our model consists of four functions, $h,g,u$ and $f$. We use for them architectures similar to the ones used in the NCP model~\citep{pakman2020}.

\subsection{Set-structured Input Generation}
\label{sec:data_generation}
\textbf{MNIST, Fashion-MNIST and ImageNet}. Each data point in our training and test data is a \textit{set} of images or image representations drawn from the original data set, without using the ground-truth labels. We apply self-labeling, as outlined below in step 4, for methods that require labeled data.
We use the training split as the data source for generating datasets during training, and the validation split for testing.

The training data is generated as follows:
\begin{itemize}
    \item[1.] A data size $N \sim (N_{\text{min}}, N_{\text{max}})$ is sampled.
    \item[2.] A clustering mixture $\bc$ is generated using the Chinese Restaurant Process (CRP), which determines the number of clusters, $K$, and the distribution of data points across each group.
    \item[3.] We then sample $K$ data points from the dataset.
    \item[4.] Instance discrimination is applied, where each cluster contains the sampled data point and a number of augmentations to the fill up each cluster size, according to the sampled mixture $\bc$. 
\end{itemize}

For these experiments, we set $(N_{\text{min}}, N_{\text{max}}) = (100, 1000)$, and choose $\alpha=1$, the hyperparameter controlling the data distribution in the CRP.

Test data is generated by setting $N=300$ for the data size, following steps 2 and 3, and then sampling original images from the dataset's categories based on the sampled mixture. 

As explained in the paper (\autoref{seq:obj_func} and \autoref{alg:training}), during training our model performs space exploration by using policies sampled from a mixture of $p_{\text{data}}(\bc|\bx)$ and 
a uniform random sample of $\bc$,
using $l \sim \text{Bernoulli}(\beta)$ to control that mixture.
\noindent For MNIST and FMNIST we set $\beta=0.99$, while for datasets with fewer images per category, such as in IN50/100/200, we set $\beta=0.999$.

\textbf{2D Mixure of Gaussians (MoG)}. We create a synthetic dataset using the following procedure for both training and testing data. Similar to the image datasets, for a fixed or sampled data size $N$, a clustering mixture is generated via the CRP, which determines the number of clusters, $K$, and the data distribution. We then sample $K$ centroids from the Normal distribution, where each centroid is generated as follows: 
    \begin{equation}
        \mu_k \sim \mathcal{N}([0]_{2\times 1}, \sigma I_{2\times 2}),
    \end{equation}
followed by 2D data-point generation based on the CRP mixture: 
    \begin{equation}
        x_i \sim \mathcal{N}(\mu_k, I_{2\times 2}),
    \end{equation}
using $\sigma = 10$.

For data size values,
we set $(N_{\text{min}}, N_{\text{max}}) = (100, 1000)$ for training and $N=300$ for testing.
Additionally, we set $\alpha=6$ for the CRP, and $\beta=0.999$ for space exploration. 

\subsection{Training Procedure}
\label{sec:training_procedure}
For both the toy-data and real-data experiments we train GFNCP with $5K$ iterations using Adam optimizer, a batch size of $64$, no weight decay and a cosine scheduler with an initial learning rate of $5\cdot 10^{-4}$ and a minimum learning rate of $1\cdot 10^{-6}$.
During training, each data point in the batch is a set-structured input, generated by following the procedure outlined in \autoref{sec:data_generation}.
During testing, we calculate the average NMI, ARI, and MC metrics (when applicable) across $M$ input sets drawn from the test data source
%, with each set containing $300$ data points. 
%
For MoG we set $M=3K$; for MNIST and Fashion-MNIST, $M$ is set to $10K$. For ImageNet, we set $M=2.5K$ due to the lower number of data points in each category.
As for the competitor's training (DAC, Set Transformer and NCP), we set the default hyper-parameters used in their published code or mentioned in their papers.

Each model was trained until convergence based to its objective function.
For each experiment we report the metrics' average and standard deviation over three different runs (using three different seeds), 
selecting the best model based on the lowest loss value.

\section{AVERAGE SCORE VS. GREEDY DECODING}
\label{sec:average_score}

In this section, we evaluate GFNCP using the average-score approach, where metrics are derived from multiple assignment samples, and compare it to the greedy-decoding approach introduced in the paper.
In the average-score approach, for each input set sampled from the test data, we generate 
assignment predictions by selecting the next assignment at each step in the sequence through sampling, instead of choosing the most probable assignment. Next, we compute the NMI/ARI metrics based on the top-100 assignments, sorted according to their predicted probabilities. We utilize the same models trained and evaluated in the Experiment section of the paper on Mixture of Gaussians (MoG), MNIST, and ImageNet-50/100/200 (IN50/IN100/IN200) datasets, maintaining the same test set size for consistency. Results are presented in~\autoref{tab:rebuttal_R3}. We observe that GFNCP is still equal or better than NCP across all datasets, and that greedy decoding yields slightly better results in most cases.

\begin{table}[hbt!]  %[ht]
    \centering
    \caption{Comparison of GFNCP and NCP using both greedy decoding and average-score approaches across the MoG, MNIST, and ImageNet-50/100/200 (IN50/IN100/IN200) datasets, with an arbitrary number of clusters.}
    \setlength{\abovetopsep}{1.3ex}
    \resizebox{0.4\linewidth}{!}{ 
    \begin{tabular}{lcccc}
         \toprule
         Dataset & \multicolumn{2}{c}{NCP} & \multicolumn{2}{c}{GFNCP} \\
                & \footnotesize{NMI} & \footnotesize{ARI} & \footnotesize{NMI} & \footnotesize{ARI} \\
         \midrule
         MoG & \textbf{0.92} & \textbf{0.89} & \textbf{0.92} & \textbf{0.89} \\[0.05cm]
         MNIST & 0.71 & 0.69 & \textbf{0.74} & \textbf{0.75} \\[0.05cm]
         IN50 & 0.53 & 0.42 & \textbf{0.60} & \textbf{0.45} \\[0.05cm]
         IN100 & 0.48 & 0.33 & \textbf{0.54} & \textbf{0.39} \\[0.05cm]
         IN200 & 0.24 & 0.15 & \textbf{0.48} & \textbf{0.28} \\
         \bottomrule
    \end{tabular}
    }
    \label{tab:rebuttal_R3}
\end{table}

\section{PREDICTING POSTERIOR FOR UNSEEN DATA}
\label{sec:online_mode}

During inference, given an input set of $N$ data points, the forward transition $P_{F}[c_n|c_{1:n-1}, x_{1:N}]$ (as formulated in~\autoref{eq:gen1}) predicts the assignment for $x_n$ based on the previously predicted assignments $c_{1:n-1}$ and the complete set of data points $x_{1:N}$.
While this is the primary formulation, GFNCP has also an "online mode", in which it is capable of predicting assignments for previously unseen data points, enabling it to perform the forward transition $P_{F}[c_n|c_{1:n-1}, x_{1:n-1}]$. This is achieved by setting $U=0$ (see~\autoref{Q}), thus eliminating the encoding of yet unlabeled points.\\

To illustrate this functional modality, we performed several experiments where we trained and evaluated GFNCP in online mode (i.e., with $U=0$) on Mixture of Gaussians (MoG), MNIST, and ImageNet-50/100/200 (IN50/IN100/IN200) datasets. In all of these experiments, we followed the exact same training procedure as described in the paper. In~\autoref{tab:rebuttal_R2} we show the performance gap between the online mode and the primary mode (reproduced from ~\autoref{tab:clustering_stats_IN_50_100_200} in the paper). 
We note that the degradation in performance is negligible.

\begin{table}[t!]  %[ht]
    \centering
    \caption{Comparison of GFNCP's online and primary modes on MoG, MNIST and ImageNet-50/100/200 (IN50/IN100/IN200), using an arbitrary number of clusters.}
    \setlength{\abovetopsep}{1.3ex}
    \resizebox{0.7\linewidth}{!}{ 
    \begin{tabular}{lcccccc}
         \toprule
         \textbf{Dataset} & \multicolumn{3}{c}{\textbf{Online Mode}} & \multicolumn{3}{c}{\textbf{Primary Mode}} \\
                & \footnotesize{NMI} & \footnotesize{ARI} & \footnotesize{MC $\downarrow$} & \footnotesize{NMI} & \footnotesize{ARI} & \footnotesize{MC $\downarrow$} \\
         \midrule
         \textbf{MoG} & 0.92 & 0.90 & 30.5 & 0.93 \footnotesize{$\pm$ 0.01} & 0.91 \footnotesize{$\pm$ 0.02} & 41.1 \footnotesize{$\pm$ 19.8} \\[0.05cm]
         \textbf{MNIST} & 0.70 & 0.66 & 33.8 & 0.72 \footnotesize{$\pm$ 0.07} & 0.74 \footnotesize{$\pm$ 0.05} & 16.4 \footnotesize{$\pm$ 9.3} \\[0.05cm]
         \textbf{IN50} & 0.59 & 0.46 & 23.9 & 0.62 \footnotesize{$\pm$ 0.08} & 0.48 \footnotesize{$\pm$ 0.06} & 60.7 \footnotesize{$\pm$ 17.8} \\[0.05cm]
         \textbf{IN100} & 0.55 & 0.40 & 54.4 & 0.60 \footnotesize{$\pm$ 0.05} & 0.40 \footnotesize{$\pm$ 0.05} & 53.0 \footnotesize{$\pm$ 13.5} \\[0.05cm]
         \textbf{IN200} & 0.62 & 0.43 & 60.0 & 0.58 \footnotesize{$\pm$ 0.04} & 0.37 \footnotesize{$\pm$ 0.02} & 41.6 \footnotesize{$\pm$ 8.2} \\
         \bottomrule
    \end{tabular}
    }
    \label{tab:rebuttal_R2}
\end{table}

\end{document}



% Supplementary material: To improve readability, you must use a single-column format for the supplementary material.
\onecolumn
\aistatstitle{Instructions for Paper Submissions to AISTATS 2024: \\
Supplementary Materials}

\section{FORMATTING INSTRUCTIONS}

To prepare a supplementary pdf file, we ask the authors to use \texttt{aistats2024.sty} as a style file and to follow the same formatting instructions as in the main paper.
The only difference is that the supplementary material must be in a \emph{single-column} format.
You can use \texttt{supplement.tex} in our starter pack as a starting point, or append the supplementary content to the main paper and split the final PDF into two separate files.

Note that reviewers are under no obligation to examine your supplementary material.